\documentclass[10pt,twocolumn,letterpaper]{article}

\usepackage{iccv}
\usepackage{times}
\usepackage{epsfig}
\usepackage{graphicx}
\usepackage{amsmath}
\usepackage{amssymb}
\usepackage{multirow}
\usepackage{adjustbox}

\usepackage[pagebackref=true,breaklinks=true,letterpaper=true,colorlinks,bookmarks=false]{hyperref}
\usepackage{bm}

\iccvfinalcopy % *** Uncomment this line for the final submission

\def\iccvPaperID{11}%4403} % *** Enter the ICCV Paper ID here

\ificcvfinal\fi
\begin{document}

%%%%%%%%% TITLE
%\title{Seeing behind Objects}
\title{Counterfactual Depth from a Single RGB Image}

\author{Theerasit Issaranon~~~~~~~~~~~~~~~~Chuhang Zou~~~~~~~~~~~~~~~~David Forsyth\\
University of Illinois at Urbana-Champaign\\
{\tt\small \{issaran1, czou4, daf\}@illinois.edu}
}

\maketitle
%\thispagestyle{empty}

%%%%%%%%% ABSTRACT
\begin{abstract}
 We describe a method that predicts, from a single RGB image, a depth map that describes the scene when a masked object is
 removed -- we call this ``counterfactual depth'' that models hidden scene geometry together with the observations.  Our method works for the same
 reason that scene completion works: the spatial structure of objects is simple.  But we offer a much
 higher resolution representation of space than current scene completion methods, as we operate at pixel-level precision
 and do not rely on a voxel representation.  Furthermore, we do not require RGBD inputs.

Our method uses a standard encoder-decoder architecture, and with a decoder modified to accept an object mask. %The network is trained with a
%combination of real and rendered images.  
We describe a small evaluation dataset that we have collected, which allows inference
about what factors affect reconstruction most strongly.  Using this dataset, we show that our depth predictions for masked objects are better than other baselines. %Given unmasked images, our approach performs comparably with the state-of-the-art single image depth prediction methods.  
\end{abstract}

%%%%%%%%% 1 INTRODUCTION
\section{Introduction}

%{\color{red}
%+ Counterfactual term\\
%+ Stereo is not helpful. Depth can't be seen by move viewer/camera by small distance.\\
%+ LayeredSceneDecomposition. -uses handcrafted loss/function to compute layered depth in contrast to neuralnetwork baseline (autoencoder). -requires high definition depth ground truth. -is slow. 
%}

%People regularly reason about free space they cannot see.
%For example, you might reach to grasp a cup, and your fingers will fold around the back of the cup, confident that there is room. % behind. %, even though you cannot see there.  
%As another example, you might put a mug down on your desk behind the laptop, even though you cannot see there.  
%While your model of this invisible space might not be precise, you have it and use it every day. 
People regularly reason about free space they cannot see. For example, you might reach to grasp a cup, and your fingers will fold around the back of the cup, confident that there is room. As another example, you might put a mug down on your desk behind the laptop, even though you cannot see there. While your model of this invisible space might not be precise, you have it and use it every day. When you do so, you are using ``counterfactual depth'' --- the depth you would see if an object had been removed.  This paper shows how to predict counterfactual depth from images.

This ability to ``see behind'' is reproduced in scene completion methods, which seek to complete voxel maps to account for the back of
objects, and to infer invisible free space.  But these methods produce limited resolution  models of space, and require
depth measurements to do so on another hand. Besides, stereo pairs provide less help to infer scene geometry behind objects, since the larger unknown depth region can't be fully observed by small changes in camera position. While there are excellent methods for inferring depth  from a single image, the resulting
depth maps represent only the free space to the nearest object. 

\begin{figure}[t]\begin{center}
    % \makebox[0.48\linewidth][c]{Input Image}\makebox[0.48\linewidth][c]{Depth Map w/o the Object}\\
    % \includegraphics[width=0.48\linewidth]{figures/1-input2.png}
    % \includegraphics[width=0.48\linewidth]{figures/1-gt.png}\\
    \includegraphics[width=1\linewidth, trim=0.7in 5.0in 4.5in 4.7in, clip]{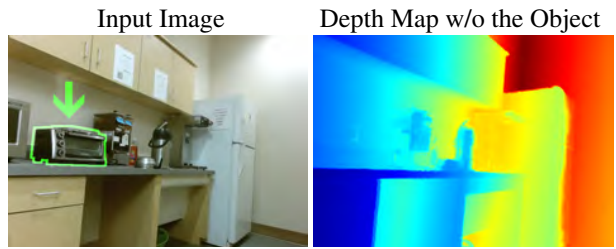}\\
    %\includegraphics[width=0.48\linewidth]{figures/1-depth.png}\\
    %\includegraphics[width=0.48\linewidth]{figures/1-output.png}
    %\makebox[0.48\linewidth][c]{Image w/o the Object}\makebox[0.48\linewidth][c]{Target Output}\\
\end{center}
\vspace{-2mm}
\caption{\textbf{Illustration.} Given an image of a scene ({\bf left}),  %and a masked object to be removed ({\bf down left}), 
our goal is to predict the depth map for that scene {\em with the object removed} ({\bf right}): \eg the image depth without the microwave (outlined in green). Our system predicts depth directly from a single RGB image, offering a representation of the free space behind an object, even though it cannot see what lies there.  These predictions are possible because indoor depth maps have quite strongly correlated spatial structure. Best viewed in color.
%Our system accepts an image of a scene {\bf top left} and predicts the depth map for that scene {\em
%    with a masked object removed} {\bf top right}.  For comparison, on the {\bf bottom left}, we show the image the
%  system {\em would have seen if the object had been removed} (but it sees the image on the top right), and on the {\bf
%    bottom right}, we show the ground truth.  Our system predicts the depth accurately, so offering a representation of
%  the free space behind an object, even though it cannot see what lies there.  These predictions are possible because
%  indoor depth maps have quite strongly correlated spatial structure. Best view in color.
}\label{fig:slash}
\vspace{-3mm}
\end{figure}

In this paper, we describe a system that can accept an
image and an object mask, and produce a depth map for the scene where the masked object has been removed~(Figure~\ref{fig:slash}): e.g. if you mask a cup in an image of a cup on a table, our system will show you the depth behind the cup. Our method works for the same reason that scene completion works.  Indoor scenes are very highly structured, and it is
quite easy to come up with very good estimates of depth in unknown regions.  However, image details are important: we
show that our method easily outperforms Poisson smoothing of the depth map.  Furthermore, our method easily outperforms
the natural baseline of inpainting the image and recovering depth from the result, because inpainting often produces unnatural pixel fields. %Code and our collected real data will be made available.

%Our system stems from the idea of generating ``counterfactual sceness'' -- real scenes with part of it been removed -- and here we directly predict on image depth to infer the occluded local scene geometry together with the global observations.
Our approach is closely related to scene completion~\cite{song2017semantic, firman2016structured}, and works for the same reason that scene completion works.  Scene geometries have quite simple spatially consistent structure. However, our method differs in important ways. We do not require additional depth information, and predict on RGB image only. Our system learns from images and depth maps (which are easy to acquire at a large scale), rather than from polyhedral 3D models of scenes. Rather than actively reconstructing the entire scene at limited resolution (voxels), our method is \textit{passive}: with no object mask, our method reports a depth map for the image; provided with a mask, it reconstructs the depth map of the image with that object removed. This deferred computation allows us to produce  representations with smoothed output and  much higher resolution than voxels can support. Our approach differs from the layered scene decomposition~\cite{liu2016layered} and depth hole filling~\cite{atapour2017depthcomp,liu2012guided} which all rely largely on the quality of input depth to perceive the hidden geometry. 

%However, our method differs in important ways: we do not produce a voxel map which is non-smooth and resolution-limited;  and Below we describe other related literature.

%Our contributions are:
%\begin{itemize}
%\vspace{-1mm}
%\item We describe a system that learns, from data, to reconstruct the depth that would be observed if an object or multiple objects were removed from a scene.
%\vspace{-1mm}
%\item For images where an object is removed, quantitative evaluations demonstrate that our method outperforms strong
%natural baselines (depth hole filling, image inpainting and then depth prediction). For images where no object is removed, the accuracy of our depth maps is comparable to that of state-of-the-art single image depth estimation methods.
%\vspace{-1mm}
%\item We introduce a carefully designed test set taken from real scenes that allows experiments investigating what scene and object properties
%tend to result in accurate reconstructions.
%\end{itemize}

Our \textbf{contributions}:
1) We describe a system that learns, from data, to reconstruct the depth that would be observed if an object or multiple objects were removed from a scene.
2) For images where an object is removed, quantitative evaluations demonstrate that our method outperforms strong natural baselines (depth hole filling, image inpainting and then depth prediction). % For images where no object is removed, the accuracy of our depth maps is comparable to that of state-of-the-art single image depth estimation methods.
3) We introduce a carefully designed test set taken from real scenes that allows experiments investigating what scene and object properties tend to result in accurate reconstructions.

\begin{figure*}[t]\begin{center}
     \includegraphics[width=1\linewidth,height=40mm]{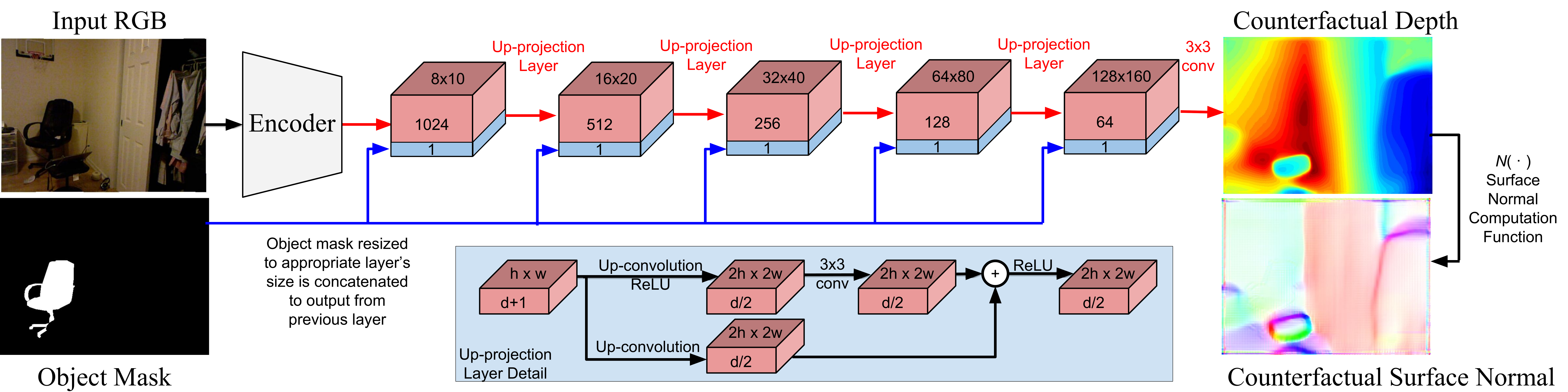}
 \end{center}
  \vspace{-2mm}
 \caption{\textbf{Network architecture.} Our network takes as input a single RGB image and a 2D object mask. The network follows an encoder and decoder strategy. The final output is the predicted depth of the scene with the object removed: we predict the depth of layouts behind the chair, and the depth of other non-removed objects, e.g. the small table in front of the chair. %Note that the network don't directly produce surface normal map, where it's computed from function $N(\cdot)$. 
 We also show the surface normal derived from our predicted depth for better illustration. Best viewed in color.}
 \label{fig:arch}
 \vspace{-2mm}
 \end{figure*}

\section {Related Work}
%https://docs.google.com/document/d/1fNHLsU8gHUKka40tmZq0HJnMhxdHjif8GIWkpsOKu1o/edit?usp=sharing

\textbf{Single image depth estimation} is now well established. Early approaches use biased models (e.g. boxes for rooms~\cite{hedau2010thinking}) or aggressive smoothing (e.g.~\cite{karsch2014depth}). Markov random field~(MRF)~\cite{saxena2006learning} and Conditional random field~(CRF)~\cite{liu2014discrete} can be applied to regress image depth against monocular images. More recent approaches use deep neural networks with multi-scale predictions~\cite{eigen2015predicting,eigen2014depth}, large-scale datasets~\cite{li2018megadepth, atapour2018real} and user interactions~\cite{ron2018monocular}.  %Chakrabarti~\etal~\cite{chakrabarti2016depth} produce depth derivatives of different orders, orientations and scales then predict a single best depth map from this overcomplete set of predictions.  
Stereo provides strong cues for unsupervised learning~\cite{godard2017unsupervised, zhou2017unsupervised} or semi-supervised learning with LiDAR~\cite{kuznietsov2017semi}. Other approaches use sparse depth samples~\cite{mal2018sparse} or variational models~\cite{kim2018deep}. Laina~\etal~\cite{laina2016deeper} propose a fully convolution approach with an encoder-decoder structure, and utilize per-pixel reverse Huber loss for better predictions. Chen~\etal~\cite{chen2016single} propose to learn from pixel pairs of relative depth, which is further improved with supervisions of surface normal~\cite{chen2017surface}. Our approach regresses on both depth and surface normal predictions. Different from Chen~\etal, we preprocess the ground truth surface normal with weighted quantized vectorization to ensure a smooth prediction. Moreover, we show in experiment that, in our task, angular-based surface normal loss can help improve performance~(while Chen~\etal found that this is less effective).

\textbf{Depth completion} helps predict the 3D layout of a scene and the objects in a novel view. The completion can be performed on point clouds~\cite{chauve2010robust}, RGBD sensors~\cite{wang2008stereoscopic,shen2013layer,barron2013intrinsic,zhang2018deep,liu2016building}, raw depth scans~\cite{pauly2005example,firman2016structured,song2017semantic} or semantic segmentations~\cite{atapour2017depthcomp}. The predictions can be represented as dense depth maps~\cite{zhang2018deep,liu2016building,barron2013intrinsic}, 3D meshes~\cite{pauly2005example,chauve2010robust}, or voxels~\cite{firman2016structured,song2017semantic}. Our ``conterfactual depth prediction'' task is challenging, because we only condition on a single RGB input and a 2D object mask only, and predict the dense depth map of the scene with the object removed -- we predict the depth that can be seen and the depth that we cannot see.

We also investigate the natural baseline of removing objects from the scene -- \textbf{image inpainting}. We can apply existing single image depth estimation approaches on the inpainted images, and obtain the predicted depth map with the objects removed. Image inpainting can be achieved by smoothing from unmasked neighbors~\cite{perez2003poisson, bertalmio2001navier, ballester2001filling}, patch-based approaches~\cite{barnes2009patchmatch, hays2007scene}, planar structure guidance~\cite{huang2014image} or convolution neural networks~\cite{iizuka2017globally,yang2017high, liu2018image,pathak2016context}. We use the method by Iizuka~\etal~\cite{iizuka2017globally}, which is one of the state-of-the-art for high resolution predictions with source code available, as our image inpainting baseline.

\section{Approach}
Assume a single RGB image $I$ is given. Now, for {\em any} object mask $M_{\text{object}}$ that identifies an object in the scene, write ${\cal M}$ for the set of pixels lying on the object.  We would like to predict the depth for the scene {\em with that object removed}~(Figure~\ref{fig:arch}). We write $d$ for the depth field; $d_{\text{behind}}$ for the depth predicted for pixels in ${\cal M}$  (i.e. the depth behind the object in the mask); and $d_{\text{observe}}$ for the depth predicted for pixels out of ${\cal M}$. For example, if the scene had a cup on a desk, and the mask lay on the cup, then $d_{\text{behind}}$ would be the desk behind the cup, $d_{\text{observe}}$ would be the rest of the desk, and $d_{\text{behind}}$ should be predictable because of the spatial coherence of objects.

\subsection{Network architecture}

Figure~\ref{fig:arch} gives an overview of our network.  We choose to modify the depth predictor by Laina~\etal~\cite{laina2016deeper}, because it is fully
convolutional, and can model the dense spatial relationship between $d_{\text{behind}}$ and $d_{\text{observe}}$.  The encoder-decoder strategy of that method
allows  coarse-to-fine corrections of $d_{\text{behind}}$. Our network's input RGB image size is $228 \times 304 \times 3$ (height $\times$ width $\times$ dimension) and the output depth map is $128 \times 160 \times 1$. The encoder is based on Resnet-50, with the fully-connected layers and the top pooling layers removed. The bottleneck feature space is $8\times10\times 1024$. The decoder consists four up-projection blocks and a $3 \times 3$ convolution layer afterwards. We use the object mask $M_{\text{object}}$ to guide the prediction by concatenating $M_{\text{object}}$ to each of the input feature layers of the up-projection block. $M_{\text{object}}$ is $0$ for pixels on the object to be removed and is otherwise $1$ for non-removed area. The bottleneck forces the decoder to capture long scale order in depth fields; the mask then informs the decoder where it should ignore image
features and extrapolate depth.  Extrapolation is helped by having an image feature encoded, because the features give some information
about the likely depth behavior at the boundary of the mask, so the decoder can extrapolate into the masked region using both depth prior statistics and feature information to guide the extrapolation. This comes at the cost of training difficulty. The decoder has a strictly more difficult task than Laina~\etal's decoder, because it must be willing to extrapolate into any masked region supplied at run time. We also experienced with concatenating the object mask with the input RGB image as input, but observed performance degrades.% alternative structures and finally conformed to this design that has the best performance.

\subsection{Network loss}\label{sec_loss} %3.2
Given a predicted image depth $\hat{d}$, and a ground truth depth $d$, the overall network loss for each image $I$ is:
\begin{align}
    L(d, \hat{d}) = &w_1 L_{\text{surface}}(d, \hat{d}) + w_2 L_{\text{avg}}(d, \hat{d})\nonumber \\
    &+w_3\text{berHu}(d, \hat{d})\label{equ:all}
\end{align}
$L(d, \hat{d})$ is the weighted summation of the surface normal loss $L_{\text{surface}}$, the average image depth difference $L_{\text{avg}}$ and the pixel-wise reverse Huber~(berHu) loss~\cite{owen2007robust}.
%Given the predicted image depth $d$, the ground truth depth $\hat{d}$ and surface normal $\hat{n}$, the overall loss  is as follows:
% \begin{align}
%     L(d, \hat{d}, \hat{n}) = &w_1 L_{\text{surface}}(N(d), \hat{n}) + w_2 L_{\text{avg}}(d, \hat{d})\nonumber \\
%     &+w_3\text{berHu}(d, \hat{d})\label{equ:all}
% \end{align}
% $L(d, \hat{d}, \hat{n})$ is the summation over the surface normal loss, the average image depth loss $L_{\text{avg}}$, and the pixel-wise reverse Huber~(berHu) loss~\cite{owen2007robust}.
%We describe the prediction of $d$ and details of each loss function in the following.
%We use the reverse Huber loss \cite{REF} to obtain stronger gradients for small errors; write
%\[B(x) = \begin{cases}|x|, &|x| \le a\\ (x^2+a^2)/2a, %&\text{otherwise} \end{cases}\]
%where $a$ is a batch dependent adaptive cutoff point.  We set $a$ as the maximum value across the absolute error of each batch. 
%We also use mean loss as described in \cite{REF}, 

%\subsection{Network Architecture}\label{method:network}%3.1

%\subsection{Surface Normal Regularization}\label{method:norm} %3.2

%\begin{figure}[t]\begin{center} %figure4A\fbox{\rule{0pt}{0.1in} \rule{0.9\linewidth}{0pt}}\end{center}\caption{Why normal loss? Graphics comparing between depth loss and normal loss. Even though both images producesmall loss on depth. On the right, normal loss is large }\label{fig:normal}\end{figure}

\textbf{Surface normal loss with weighted smoothed ground truth.} Much of the world is made of large polygons~\cite{chauve2010robust,huang2014image}, so that we can expect strong spatial correlations in surface normal.  One can obtain small depth errors with large surface normal errors, which suggests controlling surface normal error directly.
We use a loss that encourages normals derived from the predicted depth to be accurate:
\begin{align}
%L_{surface} = \frac{\sum_p\left( B\left(d_p-d_p^*\right)  - c_p \log\left(n_p \cdot n_p^*\right) \right)}{N}  + \frac{\left(\sum_p\left( d_p - d_p^* \right)\right)^2 }{2N^2}.
L_{\text{surface}}(d, \hat{d}) = - \frac{\sum_{p\in I}c_p\log\left(N(d_p)\cdot N'(\hat{d_p})\right)}{Q}\label{equ:normal}
\end{align}
$L_{\text{surface}}$ penalizes the average pixel-wise negative log likelihood of the angular distance between the predicted surface normal and the ground truth. %$\hat{n}$ denotes the ground truth surface normal.
$p$ denotes a pixel in $I$ positioned at $(x,y)$. $Q$ denotes the total number of pixels in $I$, and $c_p$ is the pixel-wise weight that we will explain later. $N'(\cdot)$ denotes the surface normal computation which is the first-order derivatives of predicted depth. %(Alg.~1 in supplemental materials). 

However, computing ground truth normals $N(\cdot)$ requires care. For two adjacent pixels with only a few millimeters apart, a small error in measurement can still produce a steep change in normal direction. %We found that a window-based gradient smoothing is crucial,  
We apply a window-based gradient smoothing method, given known camera focal length $f_x$ and $f_y$ in $x$ and $y$ dimension respectively, computing gradients $n_p=(n_{p_x}, n_{p_y}, n_{p_z})$ at pixel $p$ based on the neighboring pixels: $n_{p_x} = f_x\frac{1}{8}\sum_{i}\frac{d(x+i,y)-d(x-i,y)}{2i}$, $i\in\{1,2,\ldots,8\}$. We compute $n_{p_y}$ in the same way, set $n_{p_z} = 1$ and normalize $n_p$ to unit 1.
%We apply a window-based gradient smoothing method, first map each image pixel $p$ with ground truth depth $t=d_p$ to a 3D point $h(r,s,t)=(rt/f_x,st/f_y,t)$ given known camera focal length $f_x, f_y$ in $x$ and $y$ dimension respectively, and compute gradients $n_p=(n_{p_x}, n_{p_y}, n_{p_z})$ at pixel $p$ based on the neighboring pixels: $n_{p_x} = f_x\frac{1}{8}\sum_{i}\frac{h(r+i,s,t)-h(r-i,s,t)}{2i}$, $i\in\{1,2,\ldots,8\}$. We compute $n_{p_y}$ in the same way, set $n_{p_z} = 1$ and normalize $n_p$ to unit 1. % the computation assumes the normal in $z$ dimension is 1.
%where $d_p, d_p^*, n_p, n_p^*$ and $c_p$ are predicted depth, ground truth depth, computed normal from predicted depth, computed normal from ground truth depth, and confidence of computed ground truth's normal respectively of pixels, and $N$ is total number of pixels.

\begin{figure}[t]\begin{center} %figure4
    % \fbox{\rule{0pt}{0.1in} \rule{0.9\linewidth}{0pt}}
    \includegraphics[width=1\linewidth, trim=0.7in 4.75in 4.5in 4.65in, clip]{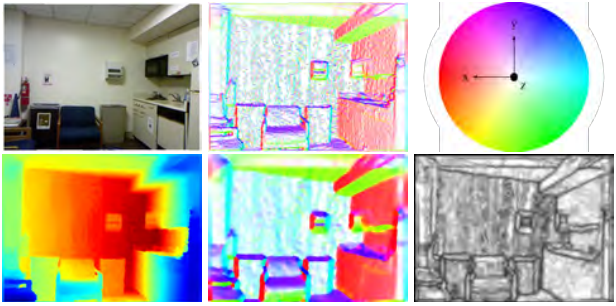}\\
\end{center}
\vspace{-2mm}
\caption{\textbf{Surface normal derived from depth v.s. our weighted quantized smoothed normal}. We show: RGB image~(top left) and the ground truth depth~(bottom left), ground truth surface normal which is the first-order derivatives of the ground truth depth~(middle top) and our weighted quantized smooth ground truth normal~(middel down). Top right is the normal direction field. Note that lighter pixel indicates that the surface normal is pointing closer to the z direction that points towards us. Bottom right is the confidence map that encourages higher confidence~(lighter pixel) for planes than boundaries. %The illustration of why we need to carefully compute normal? In the middle row, on the top is resulted normal computed directly from derivative of depth. It's subject to noise in depth acquisition. On the bottom row, we use our normal estimation to make a consistent normal for pixel in a plane along with the bottom right confidence value. Left: image and ground truth depth. Middle: normal computed by derivative of gradient and our estimation. Right: Normal direction field and confidence from our estimation.
Best viewed in color.}
\vspace{-3mm}
\label{fig:4}
\end{figure}

We then smooth the normal spatially, using a procedure to retain sharp normal discontinuities. We quantize each ground truth normal into discrete bins. We divide the hemisphere of the normal space~(assuming all pointing towards the viewpoint) into equally spanned bins of 16 latitudes and 4 azimuths. Then, we score the confidence of each bin belonging to the pixel's normal based on the weighted average angular distance to the pixel's $8 \times 8$ neighbors: $c_b = \frac{1}{64}\sum_q{(\max{(n_q\cdot n_b,0)})^\beta}$. $q$ denotes a pixel in neighborhood, $n_b$ denotes candidate bin $b$'s normal. We set $\beta=8$ to model a smooth decrease of the angle between two normal vectors going further apart. Finally, we assign the highest score to $c_p$ and its normal to $n_p$. %, the resulting ground truth is discontinuous. Instead we weight the bin normal with the bin confidence $c_b$. %In short, we compute normal of that pixel by taking into account its neighbors. Even though we lose on fine detail, fast-changing normals that occur on a rough surface or at a skinny pole, but we greatly eliminate noise. Furthermore, we use the ratio of a number of vectors in the selected bin to a number of all vectors in the averaging neighborhood to be confident value. 
The advantage of the weighting strategy is that for a flat ground truth region, most of the processed ground truth normal will be in the same bin, so we will recover a constant plane. Similarly, at a normal discontinuity (e.g. a ridge), one normal will dominate on one side and the other will dominate on the other, so the ridge will not be smoothed~(see Figure~\ref{fig:4}). We show in experiments~(Sec.~\ref{exp:obj}) that training with $L_{\text{surface}}$ helps boost our performance. %{\color{red}More algorithm detail on supplement.} %NOTE on supplement? we need to say why not use nyud-toolbox -> super slow.}
It's worth noting that our approach is faster than plane fitting~\cite{silberman2012indoor}, and is more accurate than simple partial derivatives~(please find more detailed comparison in Appendix~\ref{appx:normal}). %We see the efficiency of our smoothed ground truth normal computation, 
This is crucial since we need to re-compute surface normal for each training sample as required by the data augmentation in Sec.~\ref{method:det}. %depth scale normalization strategy requires a normal computation in each training batch, where our smoothed ground truth normal computation~(Sec.~\ref{sec_loss}) 
%~(Fig.~\ref{fig:4}). 

%This means two adjacent pixels are only a few millimeters apart and a small error in measurement, like a few millimeters, completely changes normal direction. To reduce this noise, we compute gradients from $8 \times 8$ neighborhood by solving 8th-order Taylor expansion excluding terms that change too rapid. 
%on the Surface normal to better refine the result By an assumption that world is mostly made out of plane, we should constraint of group pixel not each one individually.
%Why normal world composes of planes, When removing object, normal make depth map more consistent spatially. normal help join boundary. even though L1 is small but if boundaries are not joined it produce a large error.
%We estimate surface normal by normalizing x and y gradients scaled to proper values by camera parameters.
%But due to scaling, 1 spatial meter spans several hundred pixels.

\textbf{Depth prediction loss.} %We regress directly on the depth predictions to ensure . 
We penalize the average $\ell_2$ depth difference compared to the ground truth: $L_{\text{avg}}=\left(\frac{\sum_p d_p- \sum_p \hat{d_P}}{Q}\right)^2$. We use reverse Huber loss $\text{berHu}(d,\hat{d})$ to penalize the per-pixel prediction error, which has shown superiority in single image depth estimation~\cite{laina2016deeper}.  %Referring to Eq.~\ref{equ:all},  is a combination of $\ell_1$ and $\ell_2$ losses and . 
We set the cut-off rate $c = 0.2\max_p(|d_p-\hat{d_p}|)$ for each batch. %is the Euclidean squared distance between the mean of all predicted depth pixels and the mean of all ground truth depth pixels. 

\subsection{Implementation details}\label{method:det}

In inference, for each input image $I$ and the object mask $M_{object}$, %with labeled object to be removed, 
we first perform the largest center crop with the same aspect ratio as the network input size, then resize $I$ to fit the network input size. %We observed that preserving input images'  can help ease the network training. %$M_{object}$ is rescaled to match the various sizes of the input feature maps in the decoder layers. 
The output depth map is then resized back to the same scale as the original cropped image by bilinear interpolation. % for evaluation.

\textbf{Mask dropout.} Initial experiments indicated that depth regressions against images tend to have quite localized
support, likely because very high spatial correlations in real images mean that large-scale support is superfluous.  But
a network that predicts depth in locations where there are no known pixel values needs to have spatial support on very
long scales (so that a location where pixel values aren't known can draw from locations where the pixels are known).  To
achieve this, we randomly flip each pixel value in the object mask with a chance of 10\%, meaning a mask dropout rate of
0.1.   This forces the network to be able to use nearby pixels to predict depths.  We mask out the flipped pixels when
computing the loss to avoid error backpropagation. We show in experiments~(Sec.~\ref{exp:obj}) that training with mask dropout helps stabilize our performance. 

\textbf{Data Augmentation.} During training, we perform random cropping instead of center cropping to increase the training samples. The window size varies between the fraction $\alpha = [\frac{2}{3}, 1]$ of the size of the largest center crop. We perform the same cropping for the ground truth depth map $d$. Note that a smaller cropping is equivalent to a closer view of the object, resulting in a smaller distance to the camera. We thus divide each pixel value $d$ with $\alpha$ in order to preserve the depth scales across different crops of the same image. We also update each crop's normal given the re-scaled depth, using the weighted quantized smoothed normal computation as described in Sec.~\ref{sec_loss}. %: the per-pixel accuracy (within $20^{\circ}$) of our normal is 77\% v.s. the per-pixel accuracy by partial derivatives as 31\% evaluated by the NYUdv2 normal compuation toolbox. While our computation is faster than
%We randomly resize each input image and depth pair with a scale varies between $[1,1.5]$ and divide the depth map with the same scale to ensure the consistency correct depth proportion. Objects are larger when they are closer to a viewpoint. 
Moreover, we perform random rotation on the image plane ranges in $[-5,5]$ degrees, random horizontal flipping and
image color changes with each of the RGB channel being multiplied by the weight ranges in $[0.8,1.2]$ independently. Each augmentation parameter is
uniformly and randomly sampled from the defined range.%This also requires a normal computation in each training batch and our smoothed ground truth normal computation shows improvement.

%%%%%%%% 4 DATASET

%collected dataset and summarize the mixed dataset we use in training.

%In which decoder can easily use this information to distinguish and separate operation between real and synthesis images.

%The training data for each epoch comes from entire raw NYUDv2 training dataset and our generated Ai2thor training data.
%It is a tedious task to create a large data of real-world images with that pair and mask.
%It requires a dataset maker to capture two instances with and without an object in the scene. It also requires the maker to create a tight object mask.

\section {Dataset}\label{text:dataset} 
\subsection{Training} \label{text:traindata}

To train our method, we need triples of ground truth: RGB image, object mask, depth with masked object being removed. %where the depth has the %depth that would be seen if the
%from the image.  
Such datasets do not exist, and are difficult to make on a large scale. Instead,
we make the ground truth tuples by rendering a synthetic dataset.  %However, a rendering data may not properly represent texture or illumination. 
However, a rendered dataset may not properly represent texture or illumination. We thus combine the data with the standard NYUd v2~\cite{silberman2012indoor} real dataset (where we have only empty object masks).  Training samples are
selected uniformly across each training set (synthetic or real), with a 50\% probability of choosing one or another. We apply mask dropout on all object masks.

\textbf{Synthetic: AI2-THOR~\cite{kolve2017ai2}} is an indoor virtual environment that supports physical simulation of
objects in the scene. We modified the default simulation setting to be able to remove every object in the scene, rather than
pickupable objects only.   AI2-THOR has 120 predefined scenes from four categories of rooms:  kitchen, living room,
bedroom and bathroom. In each scene, we place an agent at a random location for 100 times. The height of agent is sampled under the normal distribution with mean of 1.0m and a standard derivation (std) of 0.1m. The agent looks
at the scene with a randomly sampled altitude, which is normally distributed with a mean of $0^o$~(looking at horizon) and a std of $10^o$. %ranges in $[-10,10]$ degree from the horizon. 
At each view, we generate the ground truth
depth map with one of the objects removed. %We perform the removal for all visible  objects. 
For each type of room, we use 27 scenes for training and withhold three scenes for testing. This creates 47k  $640 \times 480$ image-depth
pairs of synthetic samples. Each rendered depth map ranges up to 5 meters. 

\textbf{Real: NYUd v2~\cite{silberman2012indoor}} is one of the widely used RGBD dataset with real indoor scenes. We
use the official train and test split in our experiment. %We use the official inpainted depth images as our ground
%truth depth map. 

%figure5
\begin{figure}[t]\begin{center}
    % \makebox[0.32\linewidth]{NYUd v2}
    % \makebox[0.32\linewidth]{AI2-THOR}
    % \makebox[0.32\linewidth]{Ours}\\
    
    % \includegraphics[width=0.32\linewidth]{figures/5-nyu-image.png}
    % \includegraphics[width=0.32\linewidth]{figures/5-ai2-image.png}
    % \includegraphics[width=0.32\linewidth]{figures/5-our-image.png}\\
    \includegraphics[width=1\linewidth, trim=0.7in 5.1in 4.5in 5.0in, clip]{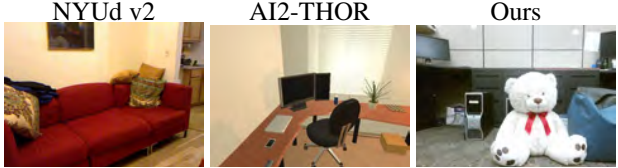}\\
\end{center}
\vspace{-3mm}
\caption{\textbf{Image samples from the dataset we use}. Left to right: NYUd v2~\cite{silberman2012indoor} (real dataset), AI2-THOR~\cite{kolve2017ai2} (synthetic dataset), our collected dataset (real dataset). AI2-THOR and our collected dataset has ground truth depth with object removed. Best viewed in color.}
\label{fig:5}
\vspace{-3mm}
\end{figure}

\subsection{Testing} \label{text:testdata} %4.3
\textbf{Synthetic}. We use the test split of AI2-THOR to compare with other baselines. We obtain 1162 test samples with depth changes of least 0.25m per pixel after the object is removed. Slight changes in depth can hardly be examined the performance. %Out of 5031 test images, 1162 images are belong to this subset where its . On the full testset only 2\% of pixels are removed on average such that it is hard to evaluated the performance. The result on the full subset and the affect of depth prediction method are in the supplemental materials.}

\begin{table}
\begin{center}
\resizebox{0.46\textwidth}{!}{
\begin{tabular}{|c|c|}
\hline
Factor & variables\\
\hline
shape complexity & simple (e.g. box), complex (e.g. chair)\\
shape rarity & common (e.g. box), rare (e.g. doll)\\
number of objects close by & 0, 1, 2\\ 
object behind & wall, empty space, other objects\\
distance to the camera &1.5m, 2.0m\\
\hline
\end{tabular}
}
\end{center}
\vspace{-1.5mm}
\caption{Factors and variables used to construct our dataset.}\label{fig:ours}
\vspace{-1.0em}
\label{tab:dataset}
\end{table}

\textbf{Real}. We have collected a small but carefully structured RGBD dataset for evaluation using Kinect v2, as shown in Figure~\ref{fig:5}.  Our dataset contains both RGB images and the depth maps before and after the removal of objects.   For each image, we carefully label a 2D
tight object mask around the object to be removed.  %when predicting depth behind the object. 
Our images are collected so as to investigate five factors that might affect the prediction error~(Table~\ref{tab:dataset}):
(1) the complexity of the object; (2) the rarity of the object in the training set; (3) number of other non-removed
objects   close by with similar depth; (4) the object location; (5) the distance between the object and the camera. The
first two  factors focus on the object itself and the latter three focus on the spatial relationship between the object
and the scene.  This results in $2\times2\times3\times3\times2=72$ testing cases. Please find more detailed dataset configurations in Appendix~\ref{appx:eval}.

\section {Experiments}

\begin{figure*}[t]\begin{center}\small
    \includegraphics[width=1\linewidth, trim=0.7in 1.9in 0.9in 1.8in, clip]{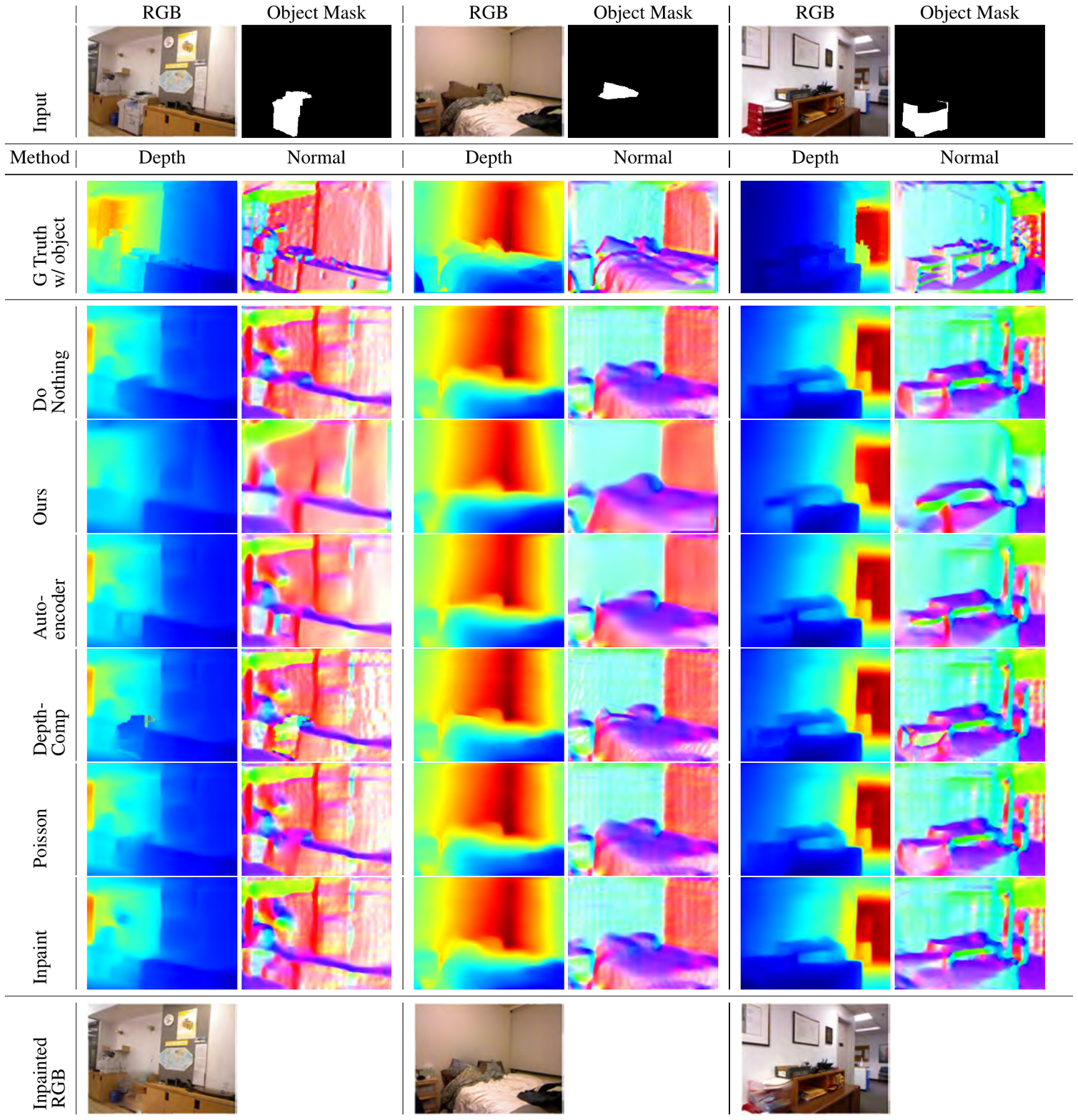}\\

\end{center}
%\vspace{-1mm}  
\caption{Qualitative results of depth estimation with the object \textit{removed} on the NYUd v2 dataset~\cite{silberman2012indoor}. We compare our approach to several baselines. %poisson depth inpainting, 
  We show in the second row the ground truth scene depth with all the object \textit{non-removed}. For image inpainting baseline we also show the inpainted RGB image for analysis. The surface normal is derived from the predicted depth. Our method is able to estimate the hidden geometry behind the cupboard when the printer is removed (column 1); the space on top of the bed when the pillow is removed (column 2); and the space below the ream of paper when the shelves but not that paper are removed (column 3). Best viewed in color.} %{\color{red}Note: There is no ground truth available after object is removed. The shown input} Best view in color.}
\vspace{-2mm}  
%Qualitative results of depth prediction with object being removed on on the NYUd v2 dataset~\cite{silberman2012indoor}. We compare our approach with two natural baselines: depth inpainting, image inpainting~(the inpainted RGB image is shown in the last column). The surface normal is computed from predicted depth. Compare with the two baselines, our method is able to predict smooth planes and clear boundaries behind the object~(the printer in the first example), and remove small objects~(pillow in the second example). Best view in color.}%Qualitative results of depth prediction with object removed . Column left to right: input RGB image and object mask, ground truth depth and normal~(object not removed), our predictions, depth inpainting, image-inpainting. The predicted normal is computed from predicted depth. We also show the inpainted image (last column). Best view in color.}
\label{fig:nyud}
\end{figure*}

\begin{figure*}[!h]\begin{center}\small
    \includegraphics[width=1\linewidth, trim=0.7in 3.8in 0.9in 3.6in, clip]{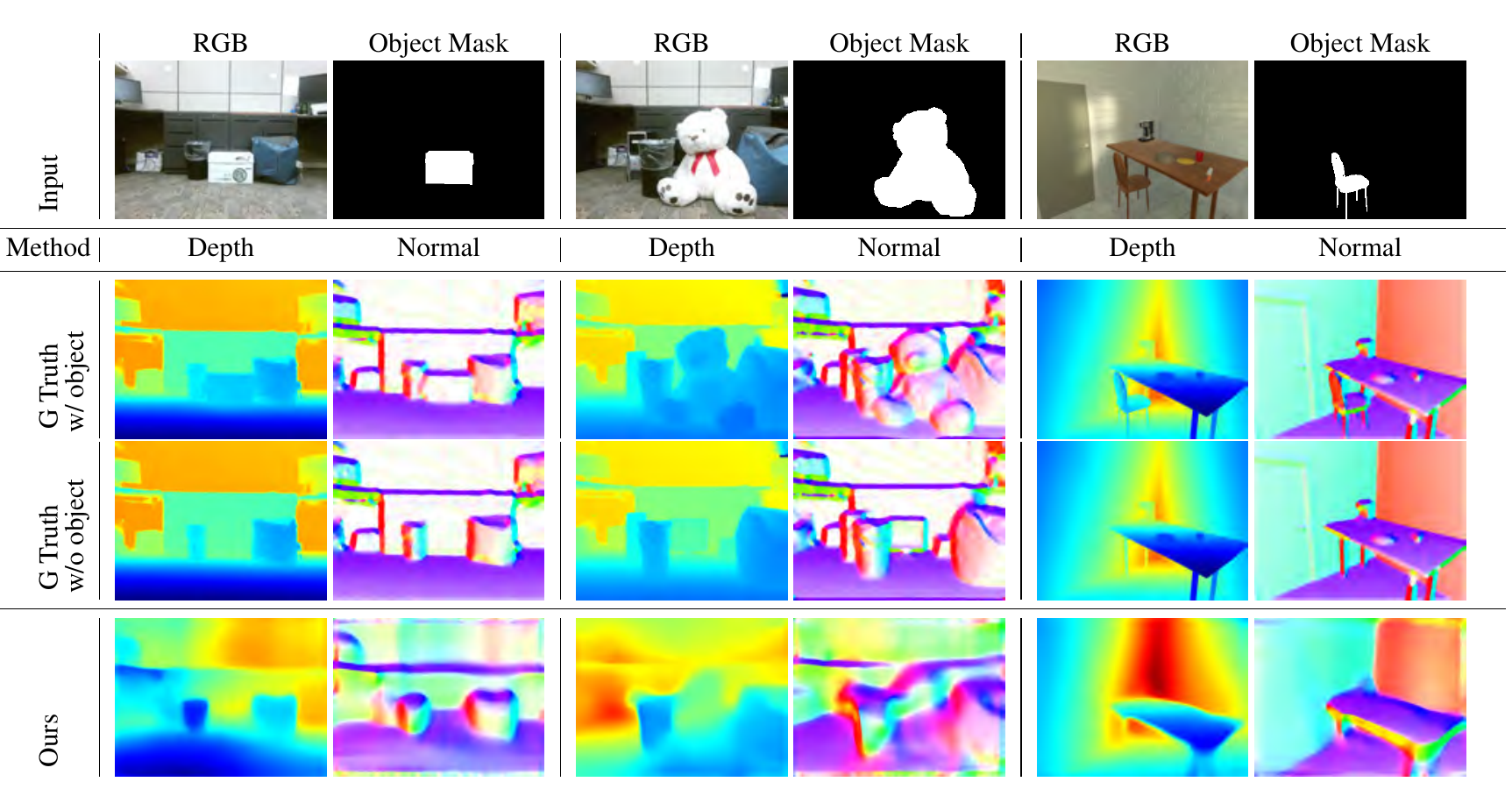}\\

\end{center}
\vspace{-3mm}  
\caption{ Qualitative results of depth estimation with the object \textit{removed} on our collected real dataset (column 1,2) and the synthetic AI2-THOR testset (column 3). %poisson depth inpainting, 
Both two datasets have the ground truth depth with the object removed shown in the third row. %by function $N(\cdot)$. 
 %where they used do-noting as and initial depth except inpainting where it is processed on inpainted image but it still use the same depth prediction as do-nothing. 
%The inpainted RGB image is shown at the last row for analysis.
Note that our method is able to predict the gap between the bin and the bag behind the center box is (column 1); the gap between the bin and the bag behind the fluffy bear (column 2); and one of the table's leg that occluded by the removed chair (column 3).  Our method has no explicit object model or semantics, and so is not puzzled by stuffed toys. Please refer to the supplemental material for comparisons with other baselines on the two datasets. Best viewed in color.}
\vspace{-1mm}  
%Qualitative results of depth prediction with object being removed on on the NYUd v2 dataset~\cite{silberman2012indoor}. We compare our approach with two natural baselines: depth inpainting, image inpainting~(the inpainted RGB image is shown in the last column). The surface normal is computed from predicted depth. Compare with the two baselines, our method is able to predict smooth planes and clear boundaries behind the object~(the printer in the first example), and remove small objects~(pillow in the second example). Best view in color.}%Qualitative results of depth prediction with object removed . Column left to right: input RGB image and object mask, ground truth depth and normal~(object not removed), our predictions, depth inpainting, image-inpainting. The predicted normal is computed from predicted depth. We also show the inpainted image (last column). Best view in color.}
\label{fig:ai2thor}
\end{figure*}

\textbf{Experimental setup.} We implement our network using MatConvNet and train it on a single NVIDIA Titan X GPU. We use the weights of pretrained ResNet-50 on ImageNet to initialize the the encoder, then train the whole network end-to-end. We use ADAM~\cite{kinga2015method} to update network parameters with a batch size of 32 and an initial learning rate of 0.01. The learning rate is then halved after every 5 epochs and the whole training procedure takes around 20 epochs to converge. In our experiment, we set the term weights in Eq.~\ref{equ:all} as: $w_1=1, w_2 = 0.5, w_3=1$ . 
%We further show how our network is able to predict depth with no object to be removed, and compare it with the state-of-the-art~(Sec.~\ref{exp:depth}). We also analyze in details on how the data distribution of the synthetic and real sets used for training can affect the performance~(Sec.~\ref{exp:distr}).
%We conduct an ablation study to demonstrate the surface normal loss improves performance~(Sec.~\ref{exp:abla}).

\textbf{Baselines.} To demonstrate the effectiveness of our approach, we compare with three classes of natural baselines:
(1) \textbf{``Do nothing''}. We simply ignore the mask and apply our approach to estimate image depth. In this case we're predicting image depth \textit{with} the object.
 %method of Laina et al to the unmasked image.  
%This checks that there is a problem to solve, but is not a strong baseline.  
(2) \textbf{Depth inpainting}. We use the object mask to remove the object from our predicted depth map, then fill in the hole using three different methods. For the first method, we apply Poisson editing~\cite{perez2003poisson} to interpolate the missing depth based on neighboring depth values. %As our results show, this is a very strong baseline.
For the second method, we apply a vanilla auto-encoder. The auto-encoder gets as input the concatenation of the depth map and the object mask, and predicts the scene depth with the object removed. The encoder~(decoder) consists five convolution layers with kernel size of $3\times 3$, with max pooling~(scale factor 2) and ReLU in between, resulting in the same $8\times 10$ bottleneck feature size as ours. We train the auto-encoder with the same setting as our approach. For the third method, we compare to the state-of-the-art depth hole filling approach DepthComp by Atapour~\etal~\cite{atapour2017depthcomp}. DepthComp requires additional input of semantic segmentation maps. We use the outputs from SegNet~\cite{badrinarayanan2017segnet} trained on SUNRGBD~\cite{song2015sun} to run the experiment. %{\color{red}LayeredScene?}
(3) \textbf{Image inpainting}. Given the object mask, we inpaint the RGB image using the method by Iizuka~\etal~\cite{iizuka2017globally}, then predict depth from the inpainted one using our approach. 

For fair comparison, we use our network with no object mask to produce the initial depth map for all baselines. %Note that the initial depth of all baselines is the predicted scene depth by our approach with no object mask. %or Laina~\etal~\cite{laina2016deeper} depending on which one has better result on depth prediction. %we use the single
%image depth prediction method by Laina~\etal~\cite{laina2016deeper} for all the  baselines. Our method uses the same backbone as this method, meaning that differences in performance are unlikely to come from the feature representation.
We evaluate the performance of our approach and all the baselines using the following standard single image depth estimation \textbf{evaluation metrics}:
\begin{itemize}
\vspace{-2.5mm}
\item \textbf{rms}: root mean squared error: $\sqrt{\frac1Q \sum_p{(d_p - \hat{d_p})^2}}$
\vspace{-2.5mm}
\item \textbf{mae}: mean absolute error: $\frac1Q \sum_p{|d_p-\hat{d_p}|}$
\vspace{-2.5mm}
\item \textbf{rel}: mean absolute relative error: $\frac1Q \sum_p{\frac{|d_p-\hat{d_p}|}{\hat{d_p}}}$
%\item DIS: something of disparity
\vspace{-2.5mm}
% \item \textbf{norm}: mean angular difference in surface normal between $N(d_p)$ and $N(\hat{d_p})$ in degree
% \vspace{-2.5mm}
\item \bm{$\delta_i$}: percentage of pixels where the ratio (or its reciprocal) between the prediction and the label is within a threshold, 1.25, to the power $i$:
    $\frac1Q \sum_p \mathbf{1}[\max{(\frac{d_p}{\hat{d_p}},\frac{\hat{d_p}}{d_p})} < 1.25^i]$. We set $i=\{1,2,3\}$.
    \vspace{-1mm}
\end{itemize}
%Below we show both quantitative and qualitative assessment of our approach. 
Note that rms, mae, and rel are error metrics~(the lower the better) and $\delta_i$ measures accuracy~(the higher the better). 
%\textbf{norm} is the surface normal metric to evaluate how consistent predicted depth is on the local scale. 
%We follow the standard depth estimation metrics as in ~\cite{laina2016deeper}. 
For detailed analysis, we calculate the average pixel performance using the metrics on the entire image~(all pixels), the region inside the mask~(interior), and the region outside the mask~(exterior). Performance on the entire image naturally shows the ability of predicting image depth with an object removed; performance on the interior region demonstrates the ability to predict the scene depth behind the object; and performance on the exterior region demonstrates the ability of predicting the depth of non-removed area.  

%{\color{red} We could not use one single set of metric to evaluate an ability to predict a depth behind an object, 'all pixels' is a good candidate but it's not emphasize on what's matter (most of pixels are not a removed object). Yet, 'interior pixels' directly demonstrates the removal ability but it ignores the preservation of the pixels beyond the remove object.} %; yet, these numbers are not represented the exterior region where the network is still needs to predict depth correctly. 
%These two metrics are computed per pixels. 
%In the perfect network, this two numbers should be comparable (where they are not significantly different from each other?) such that it should not make any significant error in or out of the object mask region. NOTE: we need to describe/give the reason in experiment part why the number in interior is lower than the exterior. This still contradicts our reasoning why we need the interioir/exterior numbers.} Please find more results in the supplemental materials.

%\input{figure-ai2thor.tex}
\begin{table*}[!h]
\begin{center}
\resizebox{\linewidth}{!}{
\begin{tabular}{|c||c|c|c||c|c|c||c|c|c||c|c|c||c|c|c||c|c|c|}
\hline
    & \multicolumn{6}{c||}{All Pixels}
    & \multicolumn{6}{c||}{Interior}
    & \multicolumn{6}{c|}{Exterior} \\
\cline{2-19}
Method
    & rms & mae & rel & $\delta_1$ & $\delta_2$ & $\delta_3$
    & rms & mae & rel & $\delta_1$ & $\delta_2$ & $\delta_3$
    & rms & mae & rel & $\delta_1$ & $\delta_2$ & $\delta_3$  \\
\hline
% Do nothing
%     &     .408 &    .164 &    15.29 &    76.0 &    95.0 &    98.9
%     &     .470 &\bf .150 &    30.16 &    74.8 &    94.4 &    98.8
%     &     .399 &    .163 &    14.10 &    76.7 &    95.4 &    99.1\\
% Inpaint% \cite{iizuka2017globally}
%     &     .455 &    .170 &    15.12 &    73.7 &    93.8 &    98.7
%     &     .515 &    .156 &    24.78 &    75.4 &    93.1 &    98.1
%     &     .447 &    .171 &    14.16 &    73.5 &    94.0 &    98.8\\
% Poisson% \cite{perez2003poisson}
%     &     .418 &    .172 &    15.77 &    74.7 &    93.7 &    97.7
%     &     .528 &    .207 &    32.74 &    66.5 &    85.5 &    90.6
%     %& \bf .399 &    .163 &    14.09 &    76.7 &    95.4 &    99.1\\
%     & * & * & * & * & * & * \\
% DepthComp%~\cite{atapour2017depthcomp}
%     &     .411 &    .165 &    15.58 &    76.0 &    95.0 &    99.0
%     &     .490 &    .166 &    32.13 &    73.7 &    94.0 &    98.0
%     %&     .399 &    .163 &    14.07 &    76.7 &    95.4 &    99.1\\
%     & * & * & * & * & * & * \\
% Autoencoder
%     &     .413 &\bf .153 &    14.68 &\bf 77.5 &\bf 95.4 &\bf 99.3
%     & \bf .400 &    .154 &    24.84 &\bf 78.4 &\bf 97.4 &    99.7
%     &     .412 &\bf .151 &\bf 13.81 &\bf 77.9 &    95.4 &\bf 99.2\\
% Ours
%     & \bf .402 &    .166 &\bf 14.62 &    76.0 &    95.4 &    99.2
%     &     .408 &    .168 &\bf 21.40 &    75.7 &    97.0 &\bf 99.7
%     &     .399 &    .164 &    13.95 &    76.5 &\bf 95.4 &    99.1\\
% \hline
Do nothing
    &     .548 &    .364 &    .158 &    75.6 &    92.5 &    98.0
    &     .667 &    .498 &    .158 &    68.6 &    92.3 &    98.8
    & \bf .539 &    .357 &\bf .156 &\bf 76.4 &    93.0 &\bf 98.2\\
\hline
Poisson% \cite{perez2003poisson}
    &     .548 &    .363 &    .158 &    75.9 &    92.6 &    97.9
    &     .691 &    .492 &    .156 &    72.2 &    92.6 &    97.3
    %&     .539 &    .357 &    .156 &    76.4 &    93.0 &\bf 98.2\\
    &*&*&*&*&*&*\\
DepthComp%~\cite{atapour2017depthcomp}
    &     .546 &    .361 &    .158 &    76.0 &    92.7 &    98.1
    &     .684 &    .490 &    .157 &    71.6 &    92.6 &    97.9
    &*&*&*&*&*&*\\
\hline
Inpaint% \cite{iizuka2017globally}
    &     .582 &    .386 &    .165 &    73.8 &    91.3 &    97.6
    &     .665 &    .479 &    .152 &    73.9 &    92.8 &    98.5
    &     .577 &    .381 &    .164 &    74.2 &    91.5 &    97.8\\
Auto-encoder
    &     .578 &    .390 &    .163 &    73.6 &    91.6 &    98.0
    &     .602 &    .441 &    .139 &    77.1 &\bf 95.3 &\bf 99.7
    &     .577 &    .388 &    .163 &    73.7 &    91.7 &    98.0\\
Ours
    & \bf .542 &\bf .359 &\bf .157 &\bf 76.3 &    92.9 &\bf 98.2
    &     .592 &    .423 &    .138 &    78.9 &\bf 95.3 &    99.4
    & \bf .539 &\bf .356 &\bf .156 &\bf 76.4 &    93.0 &\bf 98.2\\
\hline
\hline
Ours w/o mask dropout
    &     .542 &    .364 &    .162 &    75.0 &\bf 93.5 &    97.8
    & \bf .569 &\bf .407 &\bf .133 &\bf 80.2 &\bf 95.3 &    99.1
    &     .540 &    .363 &    .162 &    75.1 &\bf 93.7 &    97.8\\
Ours w/o norm
    &     .629 &    .430 &    .187 &    70.1 &    89.5 &    96.1
    &     .678 &    .490 &    .158 &    73.9 &    92.4 &    97.4
    &     .627 &    .428 &    .186 &    70.2 &    89.6 &    96.1\\
\hline
\end{tabular}
}%resizebox
\end{center}
\vspace{-2mm}
\caption{Depth estimation performance with object \textit{removed} compared with other baselines on the synthetic AI2-THOR test set. We evaluate average pixel performance on all image pixels~(All Pixels), pixels inside the object mask~(Interior) and pixels outside the object mask~(Exterior). All baselines get initial depths (without object remove) from our method with the object masked out. The ``*'' in exterior columns means that the method does not produce pixels in this region. We also show in the last two rows the ablation study of our network without mask dropout and without our surface normal loss. {\bf Bold} shows the best score in each column.%, for comparing with baselines and ablations study separately. %Bold in ablation study means it is the best. % Autoencoder performs slightly better than ours in some metrics but it trashes at pixels outside the object mask.
%Poisson and DepthComp only produce pixels within mask; that's why we report * in exterior columns where the values are equal to Do nothing. Our method outperforms baselines.
%The last block is only for an ablation study of our network. We study the effect (or lack) of mask dropout (ours w/o mask) and normal loss (ours w/o normal). They make our network (ours) more reliable on both interior and exterior regions. 
}
% To better compare with these we make a special method to Autoencoder (C\&P AE) and ours (C\&P Ours) by copy and paste the exterior from the initial depth. Thus these C\&P methods have the same exterior, hence we clearly see the result of interior pixels.
%\vspace{-1mm}
%Metric from left to right: root mean square, absolute relative difference, $\delta <1.25$, $\delta <1.25^2$, and $\delta <1.25^3$.} %Note in left columns lower is better, while higher is better in right columns.}
\label{table:depthai2thor}
\end{table*}
% \begin{table*}
% \begin{center}
% \resizebox{\linewidth}{!}{
% \begin{tabular}{|c||c|c|c||c|c|c||c|c|c||c|c|c||c|c|c||c|c|c|}
% \hline
%     & \multicolumn{6}{c||}{All Pixels}
%     & \multicolumn{6}{c||}{Interior}
%     & \multicolumn{6}{c|}{Exterior} \\
% \cline{2-19}
% Method
%     & rms & rel & norm & $\delta_1$ & $\delta_2$ & $\delta_3$
%     & rms & rel & norm & $\delta_1$ & $\delta_2$ & $\delta_3$
%     & rms & rel & norm & $\delta_1$ & $\delta_2$ & $\delta_3$  \\
% \hline
% Do nothing
%     &     .462 &&   .227 &    64.9 &    88.8 &    99.8
%     &     .480 &&    .211 &    50.8 &    87.0 &    99.6
%     &     .460 &&    .228 &    66.3 &    89.0 &    99.9\\
% Inpaint% \cite{iizuka2017globally}
%     &     .632 &&    .312 &    48.1 &    81.9 &    99.3
%     &     .633 &&    .294 &    39.7 &    90.3 &    100.0
%     &     .632 &&    .313 &    48.8 &    81.1 &    99.2\\
% Poisson% \cite{perez2003poisson}
%     &     .473 &&    .236 &    66.1 &    87.5 &    99.2
%     &     .394 &&    .178 &    73.4 &    93.8 &    99.8
%     &     .479 &&    .242 &    65.4 &    86.9 &    99.2\\
% DepthComp%~\cite{atapour2017depthcomp}
%     & && & & &
%     & && & & &
%     & && & & &\\
% Auto-encoder
%     &     .430 &&    .210 &    69.4 &    91.1 &\bf 99.9
%     &     .342 &&    .149 &    75.3 &    98.1 &    100.0
%     &     .438 &&    .216 &    68.9 &    90.5 &\bf 99.9\\
% \hline
% \end{tabular}
% }%resizebox
% \end{center}
% \end{table*}

\begin{table*}[!h]
\begin{center}
\resizebox{\linewidth}{!}{
\begin{tabular}{|c||c|c|c||c|c|c||c|c|c||c|c|c||c|c|c||c|c|c|}
\hline
    & \multicolumn{6}{c||}{All Pixels}
    & \multicolumn{6}{c||}{Interior}
    & \multicolumn{6}{c|}{Exterior} \\
\cline{2-19}
Method
    & rms & mae & rel & $\delta_1$ & $\delta_2$ & $\delta_3$
    & rms & mae & rel & $\delta_1$ & $\delta_2$ & $\delta_3$
    & rms & mae & rel & $\delta_1$ & $\delta_2$ & $\delta_3$  \\
\hline
Do Nothing
    &     .447 &    .368 &    .207 &    67.0 &    90.6 &    99.6
    &     .600 &    .513 &    .267 &    35.8 &    67.6 &    97.0
    & \bf .430 &\bf .355 &    .201 &\bf 69.9 &    92.7 &    99.8\\
\hline
Poisson% \cite{perez2003poisson}
    &     .427 &    .352 &    .198 &    69.6 &    92.8 &    99.8
    &     .394 &    .320 &    .168 &    66.8 &    93.9 &    99.9
    %& \bf .430 &\bf .355 &    .201 &\bf 69.9 &    92.7 &    99.8\\
    &*&*&*&*&*&*\\
DepthComp%~\cite{atapour2017depthcomp}
    &     .438 &    .360 &    .203 &    68.0 &    91.6 &    99.7
    &     .513 &    .424 &    .225 &    47.9 &    79.7 &    98.8
    &*&*&*&*&*&*\\
\hline
Inpaint% \cite{iizuka2017globally}
    &     .538 &    .434 &    .258 &    60.2 &    86.4 &    98.9
    &     .526 &    .445 &    .235 &    52.3 &    92.6 &    99.8
    &     .539 &    .433 &    .260 &    60.9 &    85.9 &    98.8\\
Auto-encoder
    &     .431 &    .360 &    .192 &    66.0 &\bf 95.0 &\bf 100.
    &     .353 &    .290 &    .153 &    70.5 &    97.7 &\bf 100.
    &     .437 &    .366 &    .196 &    65.5 &\bf 94.7 &\bf 100.\\
Ours
    & \bf .425 &\bf .349 &    .198 &\bf 70.6 &    93.0 &    99.8
    & \bf .310 &\bf .247 &\bf .133 &\bf 81.9 &\bf 99.6 &\bf 100.
    &     .435 &    .359 &    .204 &    69.5 &    92.4 &    99.8\\
\hline
\hline
Ours w/o mask dropout
    &     .762 &    .612 &    .272 &    38.9 &    71.3 &    90.1
    &     .517 &    .416 &    .203 &    51.3 &    87.5 &    99.3
    &     .781 &    .630 &    .279 &    37.7 &    69.7 &    89.3\\
Ours w/o norm
    &     .455 &    .364 &\bf .188 &    66.7 &    93.6 &    99.3
    &     .393 &    .310 &    .160 &    68.8 &    96.0 &    99.8
    &     .460 &    .369 &\bf .191 &    66.5 &    93.4 &    99.2\\
\hline
\end{tabular}
}%resizebox

\end{center}
\vspace{-1mm}
\caption{Depth estimation performance with object \textit{removed} compared with other baselines on our collected evaluation dataset. We evaluate average pixel performance on all image pixels~(All Pixels), pixels inside the object mask~(Interior) and pixels outside the object mask~(Exterior). All baselines get initial depths (without object removed) from our method with the object masked out.
% We evaluate average pixel performance on: all image pixels, middle: pixels inside the object mask~(interior) and pixels outside the object mask~(exterior).  Our method outperforms baselines by a substantial margin.
}
\vspace{-3mm}
%Metric from left to right: root mean square, absolute relative difference, $\delta <1.25$, $\delta <1.25^2$, and $\delta <1.25^3$.} %Note in left columns lower is better, while higher is better in right columns.}
\label{table:deptheval}
\end{table*}
\begin{figure*}[t]\begin{center}\small
\includegraphics[width=1\linewidth, trim=0.7in 3.9in 0.9in 3.9in, clip]{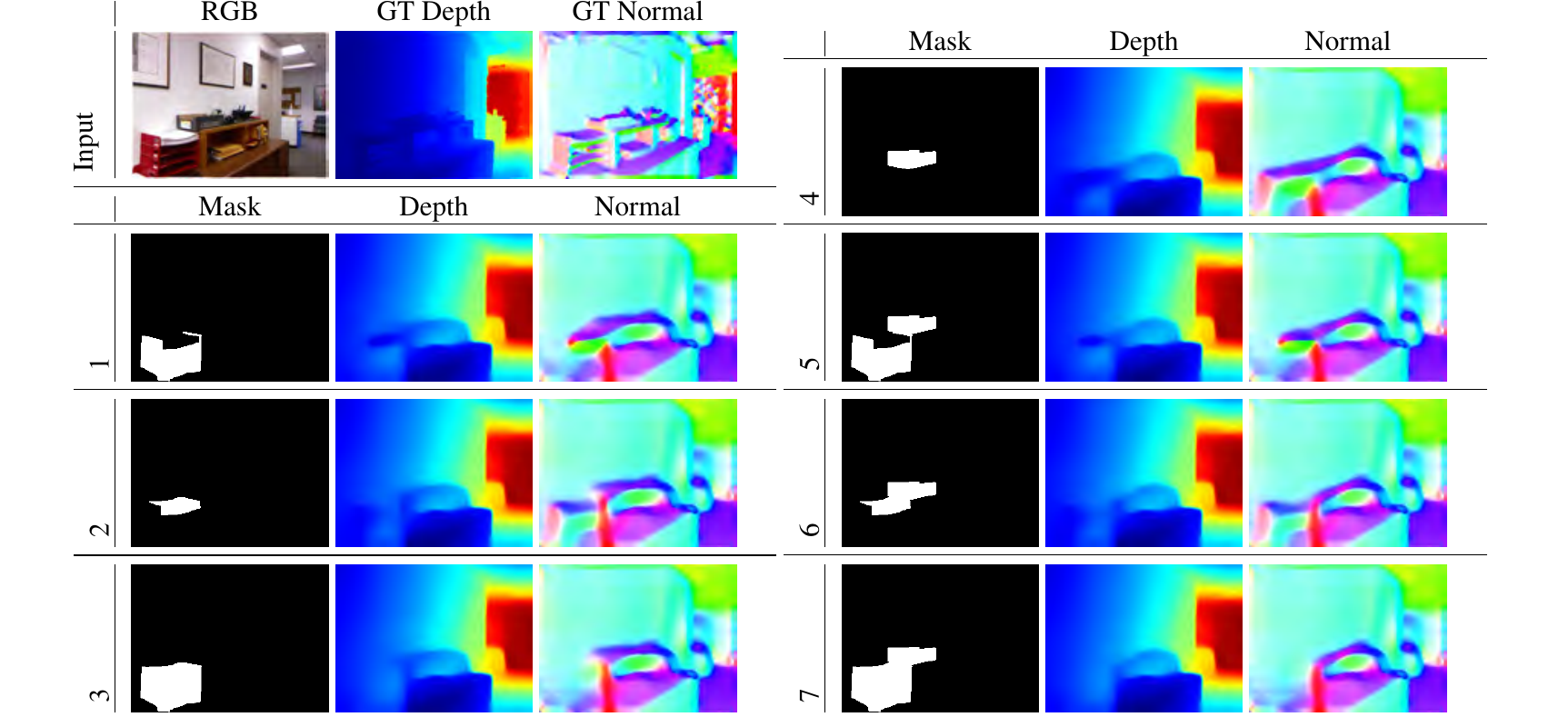}\\
\end{center}
\vspace{-5mm}  
\caption{Qualitative results of depth estimation with multiple objects \textit{removed} on the NYUd v2 dataset~\cite{silberman2012indoor}. In the first row, we show from left to right the inout RGB image, ground truth depth with all objects and the derived surface normal. In each following example we show from left to right the input object mask, our predicted depth with the object(s) removed and the derived surface normal. We show seven different input object masks as different combinations of three objects: a bookshelf, a ream of paper on the bookshelf, and the box beside the bookshelf. %We remove three object: 1 bookshelf, 2 a ream of paper on the bookshelf, and 3 the box beside the bookshelf. Object 1 is removed in example 1, 3, 5, and 7. Object 2 is removed in example 2, 3, 5, and 6. Object 2 is removed in example 4, 5, 6, and 7. 
Our network is able to remove object(s) within the supplied mask and retain other objects in the scene.
  %We show in the second row the ground truth scene depth \textit{with} the object. For image inpainting baseline we also show the inpainted RGB image~(last row) for analysis. The surface normal is computed from the predicted depth. Our method is able to estimate the depth behind the chair while preserving the depth of the person sitting on it (column 1); the depth of the flat floor after removing the chair (column 2) and the depth of the floor behind the toilet. Best viewed in color.} %{\color{red}Note: There is no ground truth available after object is removed. The shown input} Best view in color.
  }
\vspace{-4mm}  
%Qualitative results of depth prediction with object being removed on on the NYUd v2 dataset~\cite{silberman2012indoor}. We compare our approach with two natural baselines: depth inpainting, image inpainting~(the inpainted RGB image is shown in the last column). The surface normal is computed from predicted depth. Compare with the two baselines, our method is able to predict smooth planes and clear boundaries behind the object~(the printer in the first example), and remove small objects~(pillow in the second example). Best view in color.}%Qualitative results of depth prediction with object removed . Column left to right: input RGB image and object mask, ground truth depth and normal~(object not removed), our predictions, depth inpainting, image-inpainting. The predicted normal is computed from predicted depth. We also show the inpainted image (last column). Best view in color.}
\label{fig:multiobj}
\end{figure*}

\subsection {Qualitative results}

\textbf{Depth with an object removed.} We show in Figure~\ref{fig:nyud} our qualitative performance compared with other baselines on NYUd v2 dataset. NYUd v2 does not have ground truth depth with the object removed, so we could only compare qualitatively. We use the ground truth 2D segmentation in NYUd v2 as the input object mask. %We use Laina~\etal~\cite{laina2016deeper} to produce an intial depth map for all baselines.
Our approach is able to produce well-behaved depth behind the object and the depth of non-removed area, along with a good normal estimates for the hidden geometry.  Note that depth predictions by the inpainting baseline are mangled by
inpainting errors. Poisson smoothing produces somewhat better estimates, but fails in the obvious way when one side of the background is closer than the other (first column). We show in Figure~\ref{fig:ai2thor} more qualitative results on our collected real dataset and the synthetic AI2-THOR dataset.

%As Figure~\ref{fig:8} and Figure~\ref{fig:9} indicate, we can look behind objects quite successfully.  In this qualitative comparison, 
%Please refer to the normal direction field in Figure~\ref{fig:4}.

%{\color{red}
\textbf{Depth with \textit{multiple} objects removed.} One important benefit of using object mask as input is that we can arbitrarily remove any number of objects from the scene and predict the depth without these objects. Figure~\ref{fig:multiobj} demonstrates the ability of our network to estimate scene depth with different combinations of objects removed from the same scene. Our approach is also able to produce consistent predictions for non-removed area~(e.g. layouts, counter) in the same scene.
%The first object is the bookshelf is removed in examples 1, 3, 5, and 7. The ream of paper is the second object removed in examples 2, 3, 6, and 7. The last object the box beside the bookshelf is removed in examples 4, 5, 6, and 7. In this experiment it is clearly shown that our network is able to remove the specified object(s) while retaining the surrounding object.
%For example when we remove the first object, the bookshelf, the space above the bookshelf for the second object is there or not depending on the supplied mask (example 1 vs  3).
%}

\subsection{Quantitative results}\label{exp:obj} %5.4
We show in Table~\ref{table:depthai2thor} our quantitative comparison on the test set of the synthetic AI2-THOR dataset. Table~\ref{table:deptheval} reports the performance on our collected real dataset. Poisson and DepthComp do not perturb depth outside the object mask region, hence, their exterior region is equal to ``Do nothing''. We report their error metrics in exterior as *. 
%The depth prediction of an inpainted image is set to be the same as the one used as Do nothing. Therefore, these baseline methods are comparable and fair.
Our method outperforms all baselines on most metrics. Inpainting method does not work; Poisson and DepthComp have trouble removing an object. Auto-encoder and ours produce comparatively good interior (ours still slightly better) depth, but Auto-encoder produces worse depth estimates of exterior region. Note that for some measurements the depth prediction performance inside the object masked could be better than the prediction on the whole image scale. We believe that it's uncommon that objects mask other clutter, so the masked scene tends to be walls, floors, etc., where depth has simpler statistics and is easier to predict.

\textbf{Ablation study.} %{\color{red} We investigate our design choice of strategy we introduced. Details are in the supplemental materials.}
We show in Table~\ref{table:depthai2thor} and Table~\ref{table:deptheval} the performance gains by training with our smoothed ground truth normal loss (ours v.s. ours w/o normal) and the mask dropout data augmentation (ours v.s. ours w/o mask).

\textbf{Factors that affect error.} We investigate how properties of test data affect the error of the method, by regressing error against the attributes of the test images (Sec.~\ref{text:testdata}) and looking for significant predictors.  We use both individual terms and pairwise interactions, and apply an ANOVA. % For {\bf our method} the  adjusted $R^2$ is 0.882.  For the {\bf inpainting} baseline the adjusted $R^2$ is 0.633, meaning the regression is only moderate at predicting errors.
% For the {\bf Poisson} baseline the adjusted $R^2$ is 0.632.  This means that test image attributes are quite good at predicting errors for our method, but not
% for inpainting or for Poisson.  This is likely because our method is affected by whether objects are big or small, etc., but the baselines are mainly affected by simple image
% appearance properties.  In each case, relatively few interactions attain significance.  
Please find detailed analysis in Appendix~\ref{appx:anova}.

\textbf{Single image depth \textit{with} the object.} For images where no object is removed, our approach is able to predict scene depth that is of comparable quality to that of state-of-the-art single image depth estimation methods. %Another important property of our approach is . 
Please find detailed evaluations in Appendix~\ref{appx:depth}.

\section{Conclusion}

We have introduced a new task -- estimating the hidden geometry behind the object. Our method takes as input a single RGB image and an object mask, and predicts a depth map that describes the scene when the object is removed. %We describe a small real evaluation dataset that we have collected, which allows inference about what factors affect reconstruction most strongly. 
We show, both qualitatively and quantitatively, that our approach is able to predict depth behind objects better than other baselines, and is flexible in removing multiple objects. %Moreover, given unmasked images, our method shows on-par performance to the stat-of-the-art single image depth prediction approaches.
%{\color{red} We also show that our network is flexible in removing an object depending on a supplied mask.}
Our approach can be further utilized for applications like object insertion and manipulation in a single RGB image.

% One point that our approach suffered is the predicted depth lack of details. This makes removing a small object harder. On scene that has several small objects, the prediction is likely to be smooth as if no objects were there. % This makes very hard to judge the quality of the network. 
% Further improvement on the ability to resolve finer depth predictions would help improve the performance to predict depth with any size of object removed.

%EXTRA STUFF IS IN 'SUP.TEX'
%\input{sup.tex}

%\clearpage

\section*{Acknowledgements}
\vspace{-1.5mm}
This research is supported in part by ONR MURI grant N00014-16-1-2007.

{\small
\bibliographystyle{ieee}
%\bibliography{egbib}

}

\appendix
\newpage
\section{Evaluation of Surface Normal Computation}\label{appx:normal}

Due to the small magnitude of noise of the measurement error presented by the data collected from sensors in the real world~(\eg NYUd v2), it is difficult to directly calculate reliable surface normals from the depth map to train from. The error in surface normal ground truth will greatly affect the quality of the depth estimation. To incorporate $L_{\text{surface}}$ in the training procedure, we thus propose our smoothed surface normal ground truth computation in Sec.~3.2 in the main paper.
%so we try to minimized this effect as much as possible.

We demonstrate the efficacy of our surface normal computation compared to other methods on the synthetic AI2-THOR test set. AI2-THOR has accurate ground truth depth and surface normal without sensor error. We obtain the ground truth surface normal by computing the first-order derivatives from the ground truth depth. Then we simulate a measurement error by adding some random noises. We model those noises as a combination of a white noise (0.001 m) and a circle patch of diameter of 5 pixels (0.01 m) added randomly to the scene with a probability of 0.01. We shown in Table~\ref{tab:normal} the comparison between our surface normal computation, the simple gradient method and the plane fitting approach~\cite{silberman2012indoor}. ``Accuracy'' is defined as the average dot product between the computed surface normal and the ground truth~(higher the better, ranges from -1 to 1). ``Speed'' reports the time~(second) used per image. Note that our method and the gradient-based approach run on single gpu (NVIDIA Titan X) while plane-fitting runs on single cpu (1.7 GHz, 8 cores). We observed that the accuracy is highly dependent on a noise type: if we only add the random circle patch with a probability of 0.02, the resulted accuracy is 0.917, 0.831, and 0.881 respectively. 

In all, those experiments demonstrate that our surface normal computation produces high enough quality and is fast enough to be incorporated in network training.

\begin{table}
\begin{center}
\resizebox{0.44\textwidth}{!}{
\begin{tabular}{|c|c|c|c|}
\hline
& Ours & Gradient-based & plane fitting~\cite{silberman2012indoor} \\
\hline
Accuracy& \textbf{0.926}& 0.898& 0.898 \\
%Accuracy (noise 2)& \textbf{0.917}& 0.831& 0.898 \\
Speed~(s) & 0.424 & \textbf{0.013} & 78.175\\
\hline
\end{tabular}}
\end{center}
%\vspace{-1.5mm}
\caption{Ground truth surface normal computation compared with simple gradient-based approach and plane fitting~\cite{silberman2012indoor}.}\label{tab:normal}
%\vspace{-2.5mm}
\end{table}

%\section{Training Data}

\begin{figure}[t]\begin{center} %figure7
    %\includesvg[width=1\linewidth]{figures/7.svg}
    \includegraphics[width=1\linewidth,trim={0.4cm 7.1cm 0.4cm 7.1cm},clip]{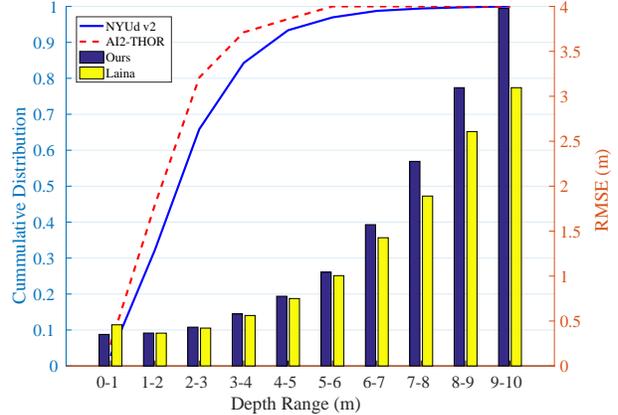}
\end{center}
\vspace{-1mm}
\caption{
Illustration of the different depth statistic between AI2-THOR and NYUd v2 dataset. The {\bf red dashed curve} shows a cumulative distribution~(cdf) of depths in the AI2-THOR dataset, which is notably biased toward
somewhat smaller depths (there are no depths greater than 5 meters).  The {\bf blue solid curve} shows a cdf for the depths in
NYUv2, which has a greater proportion of large depths.  The frequency can be read on the {\bf left} axis.  Dark bars
show the mean error in a given depth range for our method run on NYUv2; light bars show
the same for the method of Laina \etal. Note that, in the closer depth domains that are more frequent in
AI2-THOR, the two methods have comparable error; our method makes depth errors mostly on the relatively unfamiliar large
depths (and Laina's method makes depth errors on the less familiar small depths in AI2-THOR). Best view in color.
\label{histogramofdepth}
}
\end{figure}

\begin{table}
\begin{center}
\resizebox{0.99\linewidth}{!}{
\begin{tabular}{|l||c|c||c|c|c|}
\hline
Method & rms & rel & $\delta_1$ & $\delta_2$ & $\delta_3$  \\
\hline
\multicolumn{6}{|c|}{NYUd v2~\cite{silberman2012indoor}} \\
\hline
Saxana \etal \cite{saxena2006learning}
    & 1.214 & 0.349 & 44.7 & 74.5 & 89.7 \\
Eigen \etal\cite{eigen2014depth}
    & 0.877 & 0.214 & 61.4 & 88.8 & 97.2 \\
%Roy \& Todorovic \cite{roy}   & 0.744 & 0.187 & -    & -    & - \\
Eigen \& Fergus \cite{eigen2015predicting}
    & 0.641 & 0.158 & 76.9 & 95.0 & 98.8 \\
Laina \etal \cite{laina2016deeper}
    & 0.573 & 0.127 & 81.1 & 95.3 & 98.8 \\
Ma \& Karaman \cite{mal2018sparse}
    & 0.514 & 0.143 & 81.0 & \bf 95.9 & \bf 98.9 \\
Kendall \cite{kendall2017uncertainties}
    & \bf 0.506 & \bf 0.110 & \bf 81.7 & \bf 95.9 & \bf 98.9 \\
% Ours w/o normal             & 0.692 & 0.144 & 77.8 & 93.8 & 98.1 \\
Ours
    & 0.642 & 0.142 & 80.1 & 94.6 & 98.4 \\
% \hline
% \multicolumn{6}{|c|}{AI2-THOR~\cite{kolve2017ai2}} \\
% \hline
% Laina \etal \cite{laina2016deeper}
%     & 0.739 & 0.321 & 49.9 & 78.7 & 91.8 \\
% % Ours w/o normal             & 0.546 & 0.207 & 70.2 & 90.0 & 96.0 \\
% Ours
%     & \bf 0.446 & \bf 0.166 & \bf 77.2 & \bf 94.0 & \bf 98.0 \\
\hline
\multicolumn{6}{|c|}{Our Collected Evaluation dataset} \\
\hline
Laina \etal \cite{laina2016deeper}
    & 0.449 & 0.222 & 67.4 & 89.8 & \bf 99.9 \\
% Ours w/o normal             & 0.453 & \bf 0.191 & 65.7 & 93.6 & 99.3 \\
Ours
    & \bf 0.417 & \bf 0.194 & \bf 70.7 & \bf 93.0 & 99.8 \\

\hline
\end{tabular}
}
\end{center}
\vspace{-2mm}
\caption{Single image depth estimation compared with the state-of-the-art on the NYUd v2 dataset~(top) and our collected real dataset~(bottom).% on three evaluation metrics. %We evaluate on the test split of the three dataset: NYUd v2, AI2-THOR and our collected dataset. 
We directly test our trained approach that predicts depth with the object removed. In this case, our method gets an empty object mask (no object to be removed). Our approach shows on-par performance. Our approach does not obtain performance gain with more training data~(AI2-THOR) but rather gets performance drop on NYUd v2, This is because the depth distributions of the two dataset are different and training with both datasets will slightly harm performance on each other, as illustrated in Figure~\ref{histogramofdepth}. %Our method is behind the state of the art on NYUd v2, because it is trained on AI2-THOR as well as NYUd v2, and get disrupted
%frequencies of large depths (see supplement for more details). %but state of the art methods perform worse on AI2-THOR and on our evaluation dataset, because they tend to do worse than our method on nearer surfaces.%  Note that surface normal loss significantly helps our method.
\vspace{-2mm}
\label{table:depth}}
\end{table}
\section{Single Image Depth Estimation \textit{with} the Object}\label{appx:depth}
While this is not our objective, we also evaluate our performance with no object removed -- same as single image depth estimation. We directly test our trained network that predicts depth with the object removed on NYUd v2 dataset and set the input object mask as empty. %the synthetic AI2-THOR and our collected evaluation dataset. Note that other methods are trained on NYUd v2 only. %while our approach cannot be trained on NYUd v2 only as we assume a known object mask~(which can only be provided by the synthetic dataset), otherwise our method acts as the approach of Laina~\etal.
Table~\ref{table:depth} compares our method to a variety of the state-of-the-art on the NYUd v2 dataset and our collected evaluation dataset. Though our approach is trained for a different task, we show on-par performance. %Note that training with more data~(AI2-THOR) other than NYUd v2 doesn't show performance gains but rather performance drops, as our method appears to be somewhat behind Laina~\etal~(from which we derive the network architecture) on NYUd v2. This is because the depth distributions of the two datasets differ, and training with them together slightly harms the performance on each respectively.
We consider the main comparison with Laina~\etal, since we have the similar encoder-decoder structure. Our method outperforms Laina~\etal on our collected dataset but shows performance degrades on NYUd v2. We realized that this is %because after joint training with AI2-THOR synthetic data, our network performance for single image depth estimation (no object removed) drops on the NYUd v2 dataset~\cite{silberman2012indoor}, as our method appears to be somewhat behind Laina~\etal~(from which we derive the network architecture). The performance drops on depth prediction is 
due to the different depth statistic between the synthetic and real dataset. In Figure~\ref{histogramofdepth}, we show that depth maps in AI2-THOR range up to 5 meters, compared to the maximum depth of 10 meters in NYUd v2. This biases the depth predictor to be in favor of the shallower depth. As a result, our network trained on both NYUd v2 and AI2-THOR makes more significant (RMS) error on the depth prediction that is deeper than 5 meters on NYUd v2 test set in comparison to Laina~\etal~\cite{laina2016deeper}. However, it is unavoidable for us to not to use the synthesis dataset, since it is the only source that we can easily manipulate the scene to have ground truth depth with the object removed. 

In all, we conclude that our method, like others, performs very strongly on test sets where the distribution of depths compares to that in training, but degrades when it encounters novel depths.
 
%On the other hand, all experiments in Sec.~\ref{exp:obj} using our network to predict the initial depth leads to significant performance gain. Please refer to results and detailed analysis in the supplemental materials.  
%other strong published methods for the case
%Since our approach uses Laina~\etal~\cite{laina2016deeper}'s approach as our backbone for depth estimation, we compare with Laina~\etal~\cite{laina2016deeper} on the  
%We also show the reported performance by other state-of-the-art on the NYUd v2 dataset. %We show state-of-the-arts with reported performance on the NYUd v2 dataset. 
 %Qualitative comparison is in Figure~\ref{fig:6}. 
%While on NYUv2 test data  As %Figure~\ref{histogramofdepth} shows, the
%distribution of depths in AI2-THOR and NYUv2 is rather different. Laina \etal's method produces significantly weaker
%results on both our data and AI2-THOR data, likely because the method is strongly affected by the change in depth
%distributions.  %Figure~\ref{histogramofdepth} shows that the errors in our method occur mostly on large depths, which are not
%present in the AI2-THOR dataset.   Moreover, we can learn from the table that training with normal estimates produces improvements in our method.

%\input{table-sup-ai2thor.tex}
%\input{table-sup-eval.tex}

\section{Our Collected Evaluation Dataset}\label{appx:eval}

We show in Table~\ref{table:dataset_ours} the detailed configurations of our collected evaluation dataset. The configurations are based on the five factors we investigate that might affect prediction error. The top $2\times 2$ sub-table considers the object's characteristics itself: common or rare, simple or complex. The bottom sub-table, which has three rows, considers the variables of the spatial relationship with the scene: numbers of objects close by, the non-removed objects behind and the distance to the camera. We show typical samples of each of the five factors in the table.

\begin{table}[t]\begin{center} %figure4
    % \fbox{\rule{0pt}{0.1in} \rule{0.9\linewidth}{0pt}}
    \includegraphics[width=1\linewidth, trim=0.7in 3.15in 4.5in 2.95in, clip]{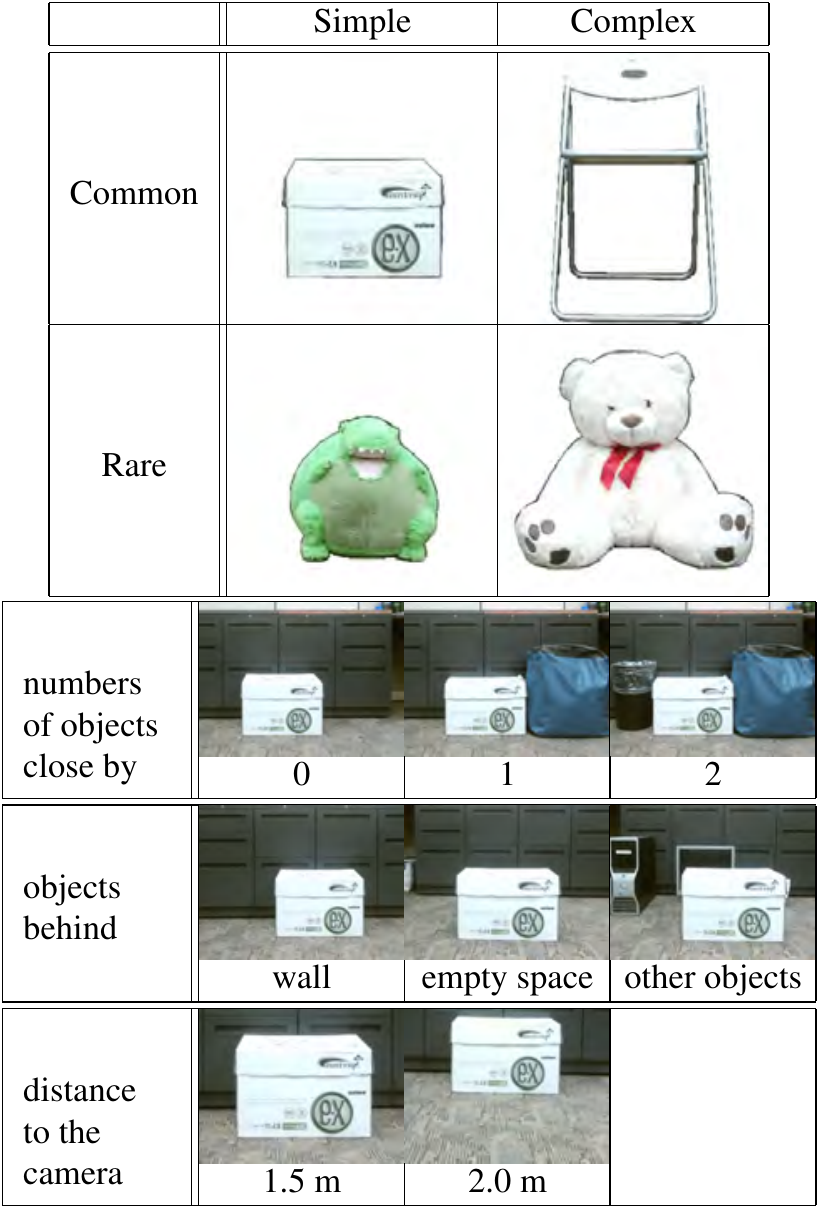}\\
\end{center}
\caption{Configurations of our collected evaluation dataset. We show sample images of each of the five factors we use to construct our dataset: object complexity, object rarity, numbers of objects close by, objects behind and distance to the camera. Note that the first two factors focus on the object itself and the latter three focus on the spatial relationship between the object and the scene. % on three evaluation metrics. %We evaluate on the test split of the three dataset: NYUd v2, AI2-THOR and our collected dataset. 
\label{table:dataset_ours}}
\end{table}

\section{More Qualitative Results}
\begin{figure*}[t]\begin{center}\small
    \includegraphics[width=1\linewidth, trim=0.7in 2.0in 0.9in 1.8in, clip,page=1]{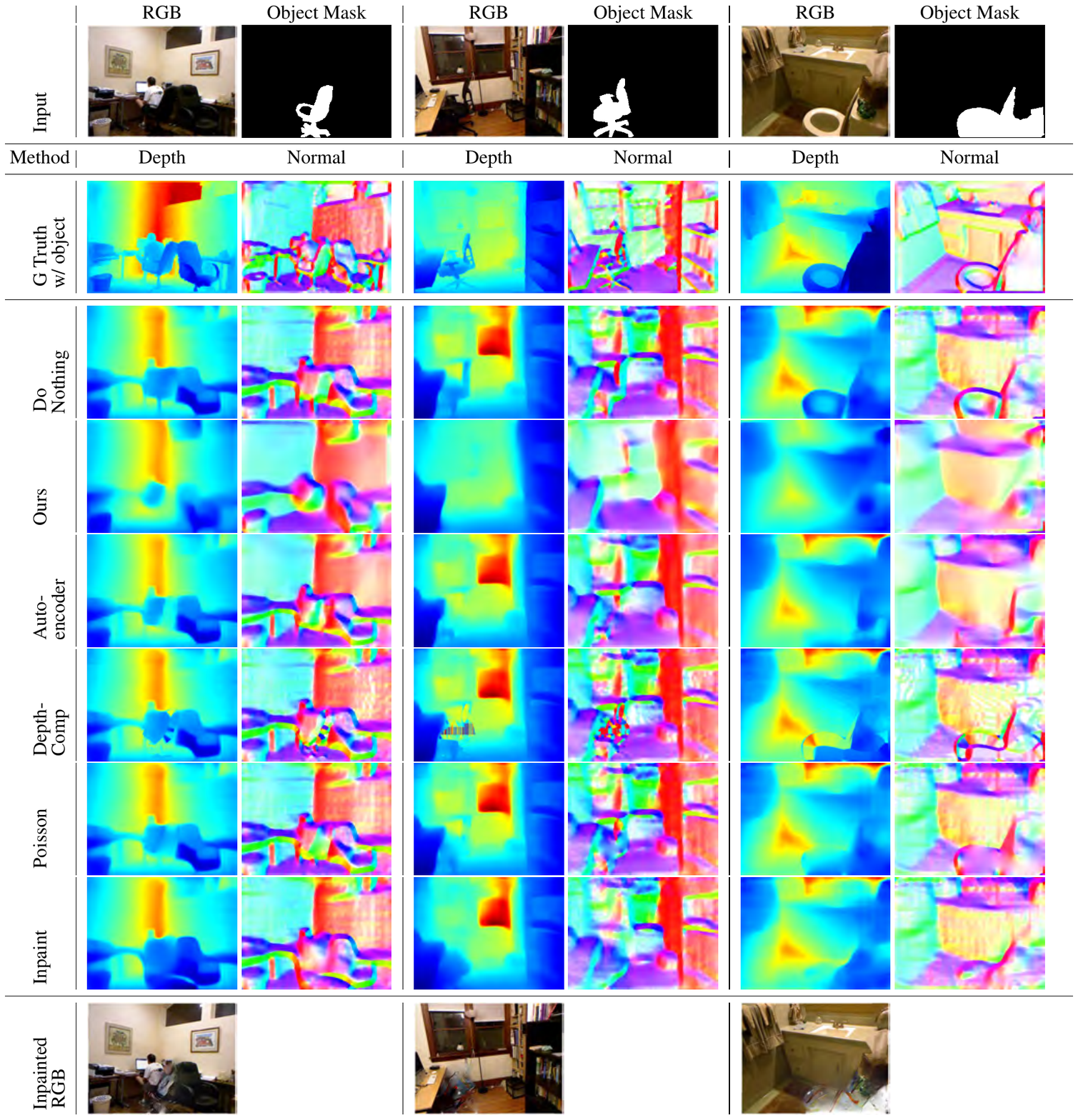}\\

\end{center}
%\vspace{-1mm}  
\caption{Qualitative results of depth estimation with object \textit{removed} on the NYUd v2 dataset~\cite{silberman2012indoor}. We compare our approach to several baselines. %poisson depth inpainting, 
  We show in the second row the ground truth scene depth \textit{with} the object.
  For image inpainting baseline we also show the inpainted RGB image~(last row) for analysis. The surface normal is derived from the predicted depth. Our method is able to estimate the depth behind the chair while preserving the depth of the person sitting on it (1st example); the depth of the flat floor after removing the chair (2nd example) and the depth of the floor behind the toilet (3rd example). Best viewed in color.} %{\color{red}Note: There is no ground truth available after object is removed. The shown input} Best view in color.}
\vspace{-3mm}  
%Qualitative results of depth prediction with object being removed on on the NYUd v2 dataset~\cite{silberman2012indoor}. We compare our approach with two natural baselines: depth inpainting, image inpainting~(the inpainted RGB image is shown in the last column). The surface normal is computed from predicted depth. Compare with the two baselines, our method is able to predict smooth planes and clear boundaries behind the object~(the printer in the first example), and remove small objects~(pillow in the second example). Best view in color.}%Qualitative results of depth prediction with object removed . Column left to right: input RGB image and object mask, ground truth depth and normal~(object not removed), our predictions, depth inpainting, image-inpainting. The predicted normal is computed from predicted depth. We also show the inpainted image (last column). Best view in color.}
\label{fig:supnyud}
\end{figure*}

\begin{figure*}[t]\begin{center}\small
    \includegraphics[width=1\linewidth, trim=0.7in 1.6in 0.9in 1.4in, clip,page=2]{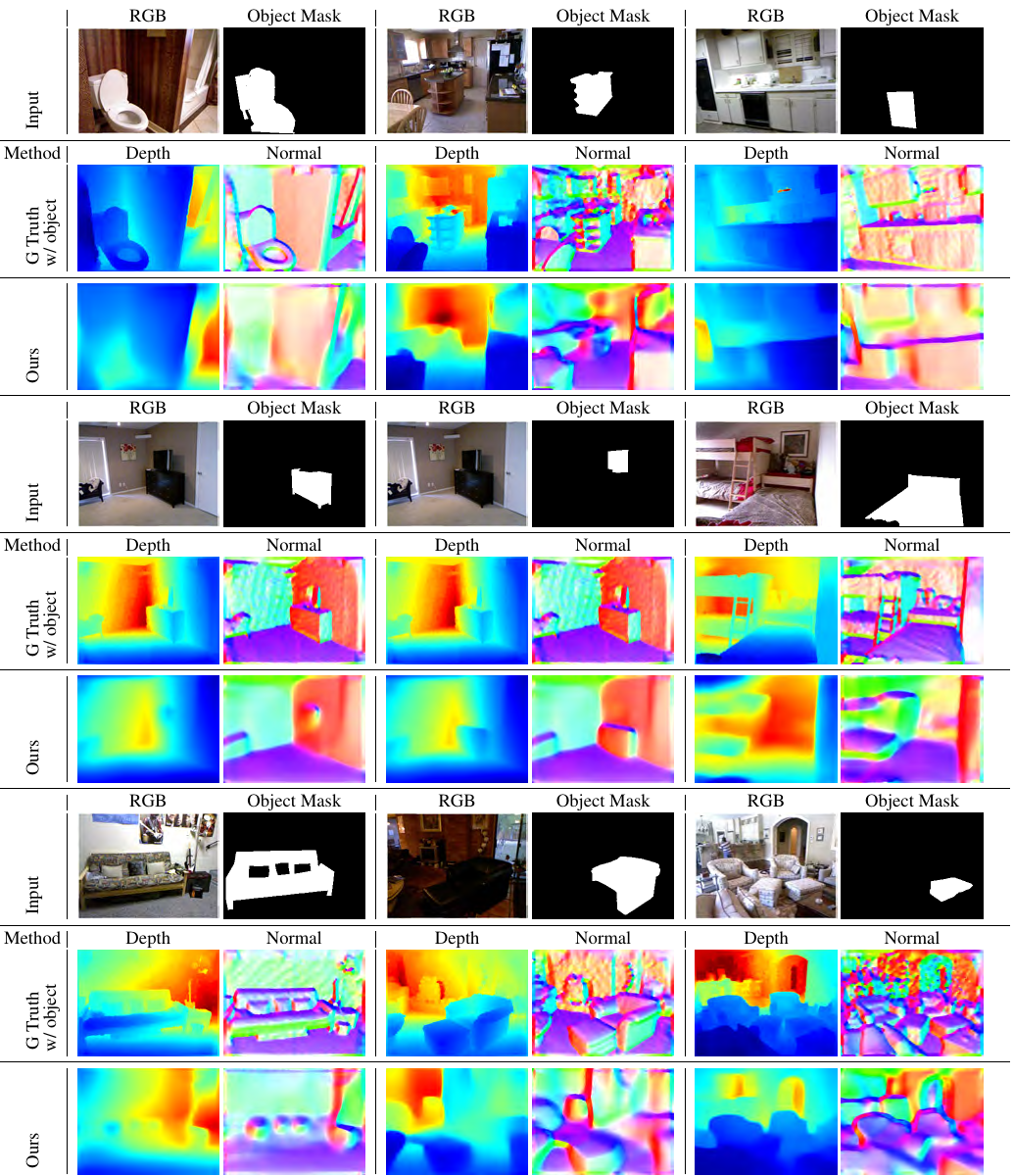}\\
\end{center}
%\vspace{-1mm}  
\caption{More qualitative results of depth estimation with the object \textit{removed} on NYUd v2 dataset~\cite{silberman2012indoor}. For each sample, we show in the first row the RGB image and the object mask, the second row the ground truth depth \textit{with} the object and the third row our predicted scene depth \textit{without} the object. Best viewed in color.%Our method can, from top to bottom, left to right, (1) recover the room layout after the toilet is removed, (2) remove the counter, (3) punch a wall to excavate the oven (note: this is an example when poisson editing ultimately fails since the surrounding depth does not change), (4) predict the corner, wall, and floor when we remove the TV stand but leave the TV intact, (5) similarly remove the TV, (6) remove the large bed and recover the flat ground, (7) remove the sofa and keep pillows afloat, (8) remove the couch behind the table, (9) remove the object in a cluttered scene.
} 
\vspace{-3mm} 
\label{fig:supnyudmore}
\end{figure*}

%In the main paper, we show the results of other baselines using our predicted scene depth as the input depth map~(with no object removed). In this supplement, we show in Fig.~\ref{fig:supnyud} and Fig.~\ref{fig:supnyudmore}, results of other baselines with the input depth map either predicted by Laina~\etal or ours~(with no object removed), compared to our predictions.
We show in Figure~\ref{fig:supnyud} and Figure~\ref{fig:supnyudmore} more qualitative results on the NYUd v2 dataet. Our method can remove objects very well. Note that our network is trained on NYUd v2 but is never trained to remove an object from this dataset~(we learn to remove an object by training on AI2-THOR). Since there is no ground truth of scene depth with an object removed in NYUd v2, we are only able to show the qualitative results compared to other baselines 

\begin{figure*}[t]\begin{center}\small
    \includegraphics[width=1\linewidth, trim=0.7in 1.6in 0.9in 1.4in, clip]{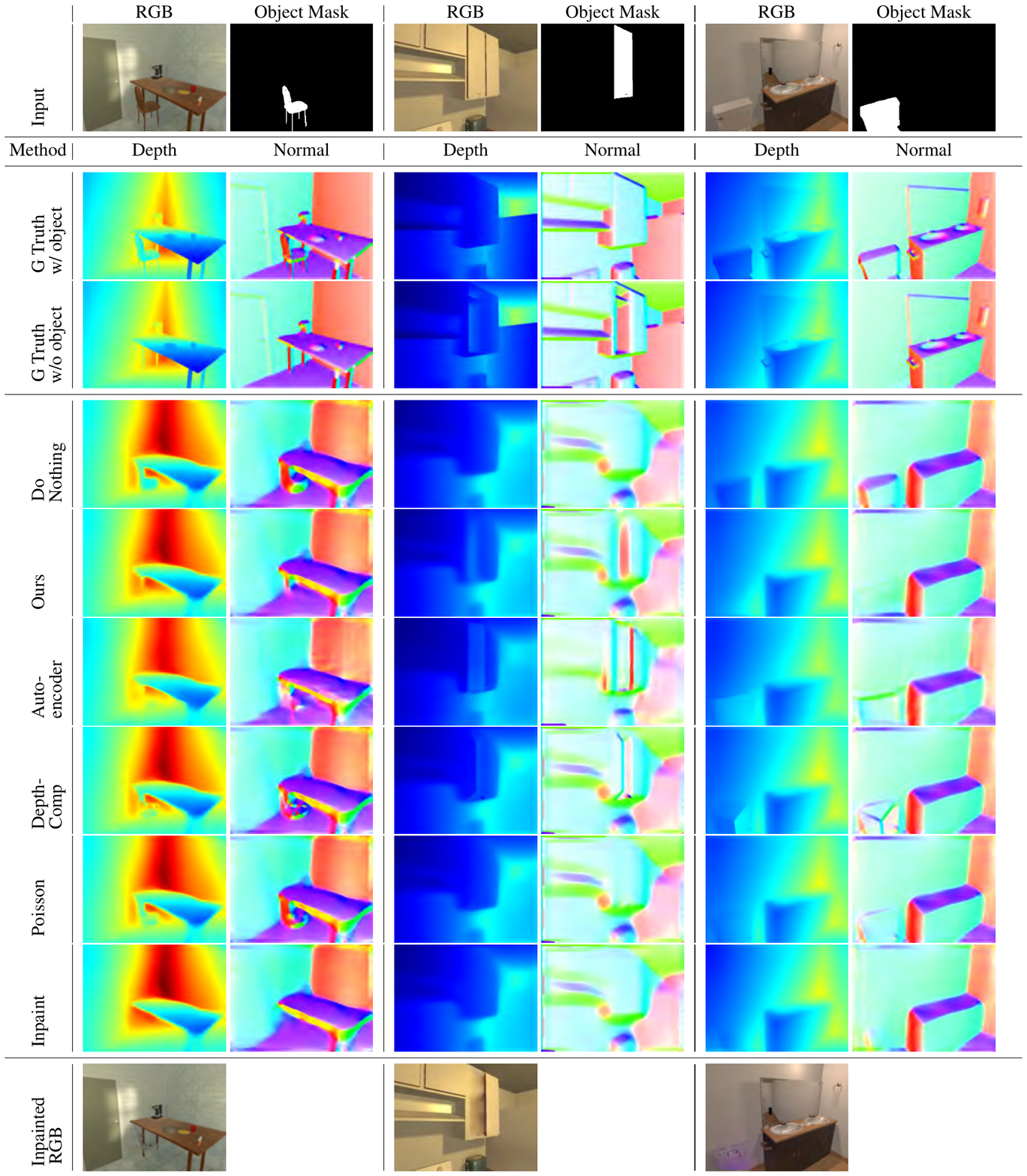}\\

\end{center}
%\vspace{-1mm}  
\caption{Qualitative results of depth estimation with the object \textit{removed} on synthetic AI2-THOR test set. We compare our approach to several baselines. We use our predicted scene depth with empty mask as the initial depth prediction for other baselines. Column 1 adds additional comparison to the sample in Figure 6 column 3 in the main paper. Note that our method could infer the space inside the cupboard (2nd example) and the wall behind the toilet (3rd example). Auto-encoder cannot fully remove the objects within the masked region: in the 2nd example, the predicted depth behind the cabinet door points to the wrong direction (as if it is added by a constant depth value); In the 3rd example, it makes a concave shape on the wall. Best viewed in color.} %{\color{red}Note: There is no ground truth available after object is removed. The shown input} Best view in color.}
\vspace{-3mm}  
%Qualitative results of depth prediction with object being removed on on the NYUd v2 dataset~\cite{silberman2012indoor}. We compare our approach with two natural baselines: depth inpainting, image inpainting~(the inpainted RGB image is shown in the last column). The surface normal is computed from predicted depth. Compare with the two baselines, our method is able to predict smooth planes and clear boundaries behind the object~(the printer in the first example), and remove small objects~(pillow in the second example). Best view in color.}%Qualitative results of depth prediction with object removed . Column left to right: input RGB image and object mask, ground truth depth and normal~(object not removed), our predictions, depth inpainting, image-inpainting. The predicted normal is computed from predicted depth. We also show the inpainted image (last column). Best view in color.}
\label{fig:supai2thor}
\end{figure*}

\begin{figure*}[t]\begin{center}\small
    \includegraphics[width=1\linewidth, trim=0.7in 2.0in 0.9in 1.8in, clip]{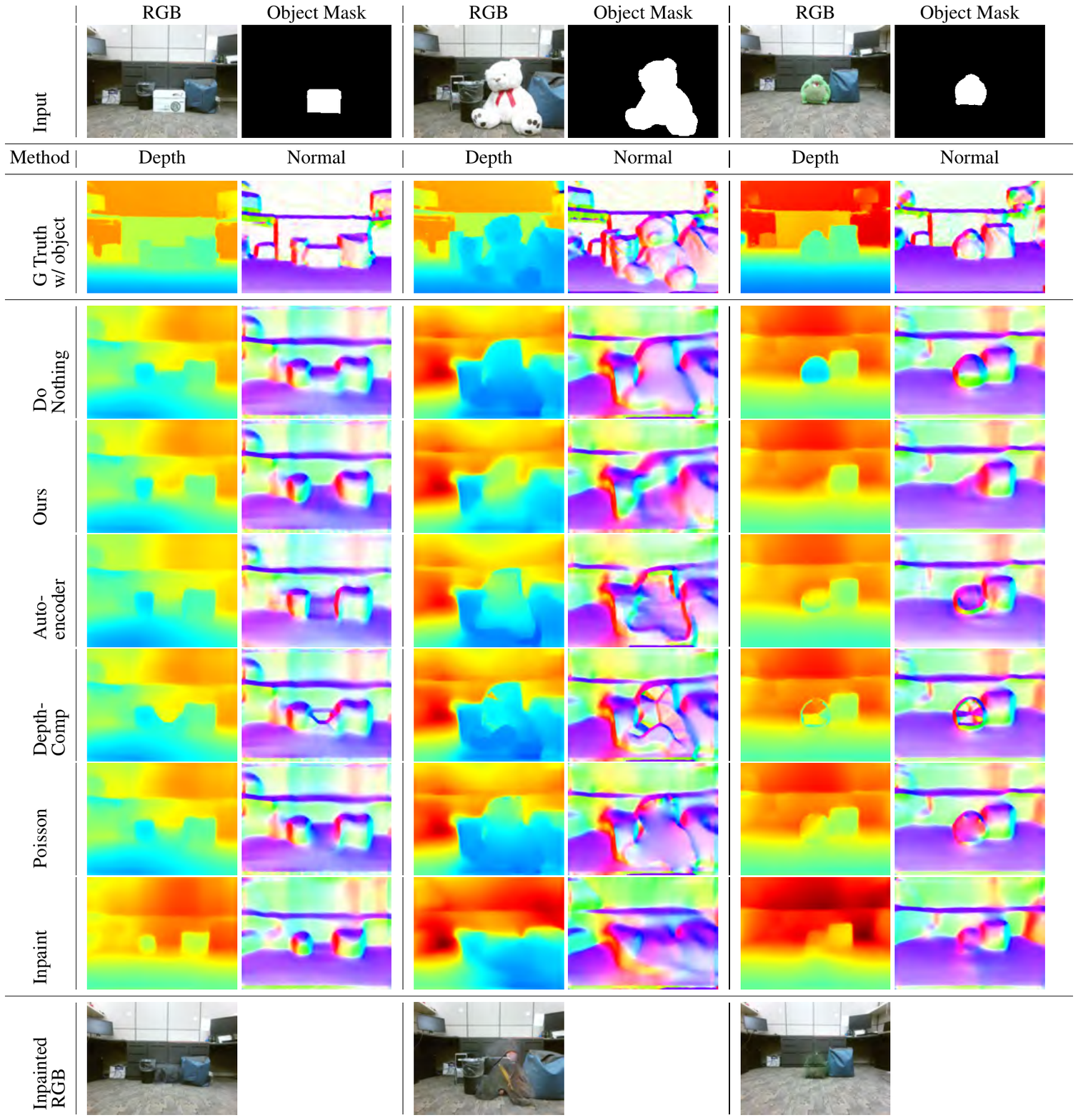}\\

\end{center}
%\vspace{-1mm}  
\caption{Qualitative results of depth estimation with the object \textit{removed} on our collected evaluation set. We compare our approach to several baselines. We use ours with empty mask as the initial depth prediction for other baselines. The first two columns adds more results to the sample in Figure 6 in the main paper. Our method is able to reconstruct the depth of the floor and wall behind the bag. Best viewed in color.} %{\color{red}Note: There is no ground truth available after object is removed. The shown input} Best view in color.}
\vspace{-3mm}  
%Qualitative results of depth prediction with object being removed on on the NYUd v2 dataset~\cite{silberman2012indoor}. We compare our approach with two natural baselines: depth inpainting, image inpainting~(the inpainted RGB image is shown in the last column). The surface normal is computed from predicted depth. Compare with the two baselines, our method is able to predict smooth planes and clear boundaries behind the object~(the printer in the first example), and remove small objects~(pillow in the second example). Best view in color.}%Qualitative results of depth prediction with object removed . Column left to right: input RGB image and object mask, ground truth depth and normal~(object not removed), our predictions, depth inpainting, image-inpainting. The predicted normal is computed from predicted depth. We also show the inpainted image (last column). Best view in color.}
\label{fig:supeval}
\end{figure*}

Figure~\ref{fig:supai2thor} and Figure~\ref{fig:supeval} shows more qualitative results of our comparison with other baselines on the synthetic AI2-THOR dataset and our collected real dataset.

%Since AI2-THOR and our collected evaluation dataset have ground truth scene depth with object removed, we show the quantitative performance in Table~\ref{table:supai2thorsel} and Tab.~\ref{table:supeval}. Our method performs the best, especially for the depth prediction in the masked region. ``Do nothing'' is slightly better in the exterior region~(outside masked region).

%In the main paper, we show the results of other baselines using our predicted scene depth as the input depth map~(with no object removed). We also use our predicted depth for other baselines in Figure~\ref{fig:supai2thor} and Figure~\ref{fig:supeval}. %In addition, we report the baseline performance using Laina~\etal as input depth map in Figure~\ref{fig:supai2thormore} and Figure~\ref{fig:supevalmore}.
%results of other baselines with the input depth map predicted by ours~(with no object removed) and Laina~\etal respectively, compared to our predictions. We show qualitative results in Fig.~\ref{fig:supai2thor}, Fig.~\ref{fig:supai2thormore}, Fig.~\ref{fig:supeval} and \ref{fig:supevalmore}. Also note that baselines use a predicted depth from Laina \etal as a starting point on the upper half of Tab.~\ref{table:supai2thorsel} and Tab.~\ref{table:supeval}, and in Fig.~\ref{fig:supai2thormore} and Fig.~\ref{fig:supevalmore}.

\clearpage

\clearpage

\clearpage

\clearpage

\section{Analysis of Variance~(ANOVA)}\label{appx:anova}
We investigate how properties of test data affect the error of the method, by regressing error against the attributes %(rows in Table~\ref{table:dataset_ours})
of the test images from our collected dataset and looking for significant predictors.  We use both individual terms and pairwise interactions, and apply an ANOVA.  %For {\bf our method}
%the  adjusted $R^2$ is 0.882.  For the {\bf inpainting} baseline the adjusted $R^2$ is 0.633, meaning the regression is only moderate at predicting errors.
%For the {\bf Poisson} baseline the adjusted $R^2$ is 0.632.  This means that test image attributes are quite good at predicting errors for our method, but not
%for inpainting or for Poisson.  This is likely because our method is affected by whether objects are big or small, etc., but the baselines are mainly affected by simple image
%appearance properties.  In each case, relatively few interactions attain significance.  
We consider the following five \textbf{individual terms}:
\begin{enumerate}
\vspace{-2.5mm}
\item object complexity
\vspace{-2.5mm}
\item object rarity
\vspace{-2.5mm}
\item numbers of objects close by
\vspace{-2.5mm}
\item background (objects) behind
\vspace{-2.5mm}
\item object's distance to the camera
\end{enumerate}

The \textbf{interaction terms} are the 2-combination of the five individual terms, resulting in ${5\choose 2} = 10$ terms.

We analyze on our approach and the two baselines: image inpainting and poisson inpainting.

\subsection{Our method}

For our method, only 3 of 10 interaction terms achieve significance using the usual F-test (i.e. $p<0.05$).   The adjusted $R^2$ is 0.882,
meaning the regression is quite good at predicting errors, and so it is reasonable to infer hard cases from regression
coefficients. The significant cases are:
\begin{enumerate}
\vspace{-2mm}
\item  objects {\tt far} from the camera with {\tt cluttered} backgrounds (mild increase in error rate); 
\vspace{-2mm}
\item {\tt simple} objects that are {\tt far} from the camera (mild increase in error rate);
\vspace{-2mm}
\item {\tt simple} objects that are {\tt rare} (mild increase in error rate). 
\vspace{-2mm}
\end{enumerate}
Of the individual terms, rarity, objects behind and distance to the camera have effects, with {\tt common} objects, {\tt cluttered} or {\tt empty space} behind, and objects {\tt far} from the camera are each associated with an increase in error rate. It is odd that {\tt common} objects should be associated with increased error, and it is odd that  objects {\tt far} from the camera should be associated with increased error.  The effect of objects behind is easily understood: the cases are {\tt against  wall}, on {\tt cluttered} or on {\tt empty} background, and it is relatively natural that predicting the depth to a wall an object is in contact with might be more accurate.

\subsection{Image inpainting baseline}
For the image inpainting baseline, again only 3 of 10 interaction terms achieve
significance using the usual F-test (i.e. $p<0.05$).  The adjusted $R^2$ is 0.633,
meaning the regression is only moderate at predicting errors.  This is likely because the conditions we investigate have
only mild effect on whether inpainting is likely to be successful (more important is image appearance around the object).
Significant effects are: 
\begin{enumerate}
\vspace{-2mm}
    \item  objects {\tt far} from the camera with {\tt two other objects close by} (mild decrease in error rate);
    \vspace{-3mm}
    \item {\tt simple} objects that are {\tt far} from the camera (mild decrease in error rate);
    \vspace{-2mm}
    \item {\tt rare} objects that are {\tt far} from the camera (mild decrease in error rate).
    \vspace{-2mm}
\end{enumerate}
 
Of the individual terms, complexity, rarity, objects behind and size have
effects, with {\tt simple} objects, {\tt common} objects, {\tt
clutter}ed or {\tt empty} backgrounds, and objects {\tt far} from the camera are
each associated with an  increase in error rate.

\subsection{Poisson editing baseline}
For the Poisson baseline, again only 3 of 10 interaction terms achieve
significance using the usual F-test (i.e. $p<0.05$).  The adjusted $R^2$ is 0.632,
meaning the regression is only moderate at predicting errors.  This is likely because the conditions we investigate have
only mild effect on whether smoothing is likely to be successful (more important is the pool of depths around the object).
Significant effects are:
\begin{enumerate}
\vspace{-2mm}
    \item  objects {\tt far} from the camera with {\tt two other objects close by} (mild decrease in error rate);
    \vspace{-3mm}
    \item {\tt simple} objects that are {\tt far} from the camera (mild decrease in error rate);
    \vspace{-2mm}
    \item {\tt rare} objects that are {\tt far} from the camera (mild decrease in error rate). 
    \vspace{-2mm}
\end{enumerate} 
Note that the above effects are the same as for the inpainting baseline.
Of the individual terms, complexity, rarity, objects behind and size have
effects, with {\tt simple} objects, {\tt common} objects, {\tt
clutter}ed or {\tt empty} backgrounds, and  objects {\tt far} from the camera are
each associated with an  increase in error rate (again, the same as the inpainting baseline).

\end{document}